\documentclass[runningheads]{llncs}

\usepackage{eccv}
 
\usepackage{eccvabbrv}

\usepackage{graphicx}
\usepackage{booktabs}
\usepackage{url}
\usepackage{epsfig}
\usepackage{amsmath}
\usepackage{amssymb}
\usepackage{caption}
\usepackage{multirow}
\usepackage{url}
\usepackage{xcolor}
\usepackage{xspace}
\usepackage{todonotes}
\usepackage{pifont}
\usepackage{wrapfig,lipsum,booktabs}
\usepackage[accsupp]{axessibility}  
\usepackage{makecell}
\usepackage{svg}
\usepackage{amsmath} %

\usepackage[breaklinks,colorlinks,citecolor=eccvblue]{hyperref}
\usepackage{orcidlink}

\begin{document}

\newcommand{\methodname}{LUNA\xspace}
\title{\methodname: Learning Universal 3D Human Animation Beyond Skinning}

\titlerunning{LUNA}
\author{Peng Li\inst{1}\thanks{Work was done during an internship at Meta.} \and
Rawal Khirodkar\inst{2} \and
Junxuan Li\inst{2} \and
Yuan Dong\inst{2} \and
Chen Cao\inst{2} \and
Yuan Liu\inst{1} \and
Wenhan Luo\inst{1} \and
Yike Guo\inst{1} \and
Shunsuke Saito\inst{2}}
\authorrunning{P. Li et al.}
\institute{The Hong Kong University of Science and Technology \and
Codec Avatars Lab, Meta}

\maketitle

\begin{figure}
\vspace{-2em}
    \centering
    \includegraphics[width=\linewidth]{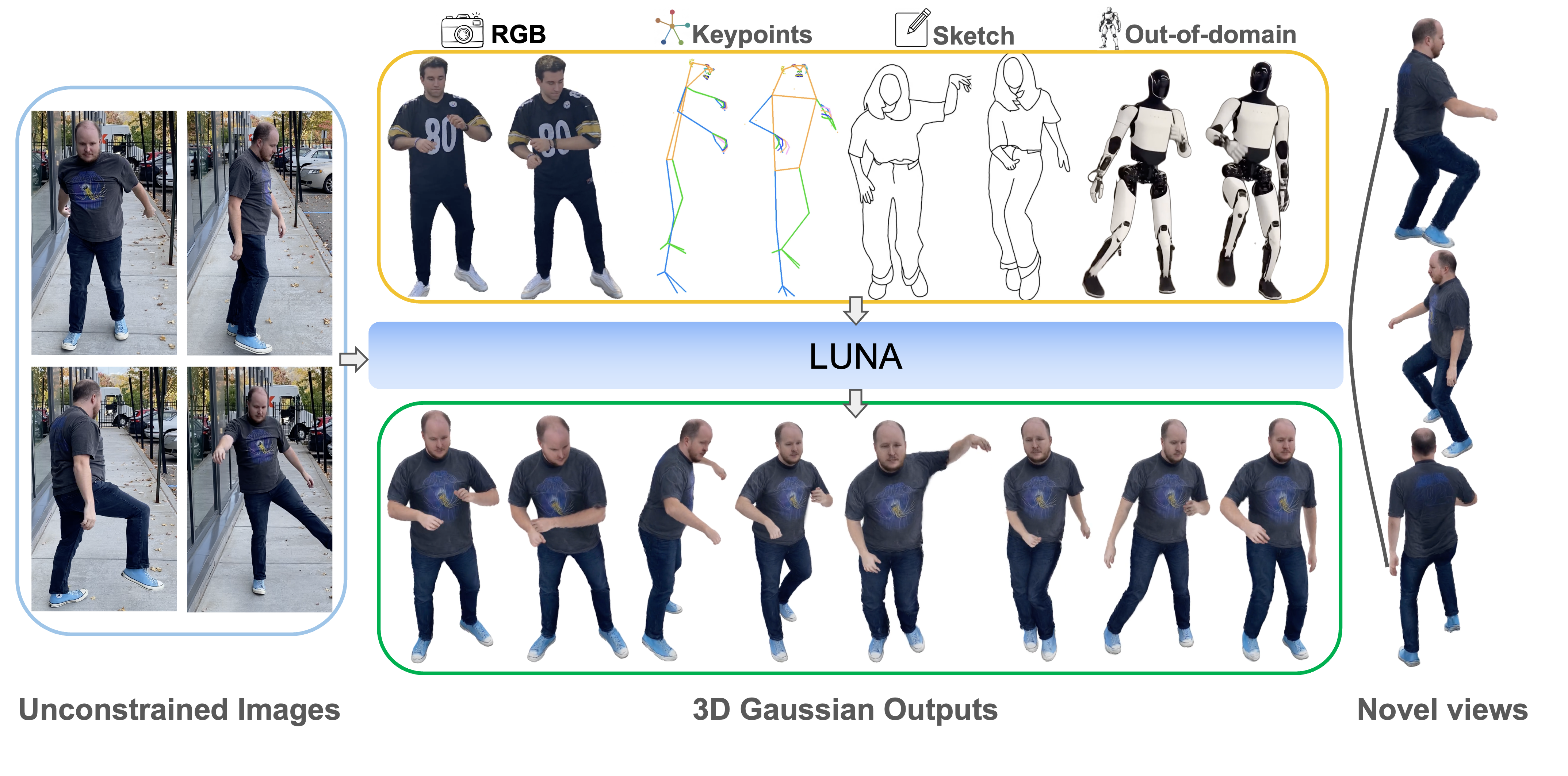}
    \vspace{-0.3in}  
    \captionof{figure}{
    Given a handful of human images, LUNA reconstructs a high-fidelity animatable 3D avatar, supporting versatile 2D control signals - including RGB images, 2D keypoints, hand-drawn sketches, and other unseen characters - without any additional preprocessing. Project page: \url{https://penghtyx.github.io/LUNA/}.
    } 
    \label{fig:teaser}
\vspace{-0.2in}  
\end{figure}

\begin{abstract}
\vspace{-0.2in}  

\vspace{-10pt}

Creating photorealistic, animatable 3D human avatars from monocular images still largely depends on Linear Blend Skinning (LBS) and parametric body models which constrain expressivity and often introduce artifacts due to imperfect fitting. We propose \methodname, an LBS-free universal neural animation model that directly maps multiple 2D controls like images, keypoints, sketch and unseen characters into 3D Gaussian deformations, bypassing explicit body fitting. 
At its core, a transformer-based motion regressor disentangles global rigid motion from fine-grained local dynamics to capture both coherent movement and subtle non-rigid effects. To resolve the inherent ambiguity of 2D-to-3D lifting while scaling beyond fitted datasets, we introduce hybrid supervision that distills soft structural priors from an LBS teacher and a loss that supports training on both limited fitted data and large in-the-wild unlabeled videos. Extensive experiments show \methodname achieves competitive visual fidelity compared to LBS-based approaches, while delivering realistic human motion and zero-shot cross-identity generalization across diverse driving modalities. To the best of our knowledge, \methodname is the first end-to-end 3D animatable model that supports implicit 2D driving.



\end{abstract}

\section{Introduction}
\label{sec:intro}

Creating photorealistic, fully animatable 3D human avatars from monocular inputs has long stood as a central ambition at the intersection of computer vision and graphics, driven by immense demand in film production, interactive entertainment, and VR/AR. Traditional methods~\cite{bagautdinov2021driving, li2024animatable} can deliver high fidelity, but they typically depend on costly multi-view capture and iterative optimization, confining scalability to controlled studio settings. In contrast, most monocular animation systems heavily rely on predefined parametric body models~\cite{loper2015smpl,smplx:2019,MHR:2025}, a convenient prior that nonetheless constrains expressiveness and often compromises photorealism under complex, highly dynamic motion. 

\vspace{1mm}\noindent
Large Reconstruction Models (LRMs)~\cite{hong2023lrm} have recently demonstrated striking generalization in 3D reconstruction, enabled by scalable transformer architectures and large datasets. Following this, several methods~\cite{qiu2025lhm, zhuang2024idol, guo2025vid2avatar} predict canonical 3D Gaussians in a feed-forward manner while incorporating parametric priors for animatability. Despite their efficiency and generalization, the continued dependence on parametric models creates a persistent representation bottleneck. First, during training, these methods commonly use Linear Blend Skinning (LBS) to map a learnable canonical representation into posed space for rendering supervision. Consequently, inevitable body fitting errors lead to suboptimal canonical learning. Second, at inference, monocular 3D pose estimation is inherently ill-posed, and small joint errors can readily amplify into temporal jitter or geometric artifacts in the rendered avatar. 

\vspace{1mm}\noindent
In this paper, we introduce \textbf{LUNA}, an LBS-free Neural Animation model that enables direct 3D animation from flexible 2D driving signals, including RGB images, sketches, and 2D keypoints. The central idea is to remove the parametric-body-bottleneck by learning a direct 2D-to-3D deformation mapping. Concretely, we propose a transformer based \emph{disentangled neural animator} that predicts deformations in two parts: a global rigid motion that accounds for coarse pose and camera-aligned movement, and a local non-rigid field that captured pose-dependent dynamics at the level of individual Gaussians. By leveraging the long-range context of our model, \methodname extracts motion cues from the 2D input and translates them into per-Gaussian deformations. Our end-to-end setup allows the model to handle large articulated motion while preserving fine-grained, non-rigid effects that are difficult to express with LBS.

\vspace{1mm}\noindent
Directly learning 3D deformation from 2D inputs using only rendering supervision, however, is severly underconstrained: without an explicit notion of human shape, reconstructions can collapse into flattened configurations, see Fig.~\ref{fig:abla_distill_mvfinetuning}.  We therefore introduce a hybrid supervision strategy. While \methodname is designed to surpass the limitations of LBS, we retain its utility as a \emph{soft} structural prior. Concretely, we utilize an LBS-based teacher model not as a hard driver, but as a regularizer that distills structural cues into the student. Combined with photometric rendering losses, this distillation preserves geometric integrity without inheriting the teacher's representational limits. Finally, to scale beyond scarce body-tracked data, we use a loss reweighting scheme that dynamically balances rendering and distillation objectives based on the proportion of labeled data within each batch. This enables our model to scale efficiently from limited annotated samples to massive, unlabeled in-the-wild video collections.

\vspace{1mm}\noindent
A notable outcome of this formulation is the emergence of \textbf{cross-identity driving}. Although trained with matched source and driving identities, \methodname learns a deformation field that captures universal humanoid kinematics. At inference, it can be driven across identities implicitly by heterogeneous modalities - keypoints, sketches, or out-of-domain character images—without further tuning. Our results demonstrate that implicit driving using \methodname drastically reduces pose-induced jitter and better captures non-rigid dynamics than parametric baselines. Our contributions are summarized as follows:
\begin{itemize}
    \item To the best of our knowledge, \methodname is the first model to enable universal 3D human animation directly driven by versatile 2D signals, establishing a new paradigm that bypasses explicit 3D control.

    \vspace{1mm}
    \item We propose a body-fitting-free animation setup that extracts motion semantics directly from 2D inputs without any preprocessing~(e.g., body fitting or foreground segmentation), thereby reducing error accumulation and significantly mitigating temporal jitter.

    \vspace{1mm}
    \item We introduce a disentangled neural animator with a fully LBS-free decoder, enabling direct modeling of pose-dependent deformations from coarse articulations to complex non-rigid dynamics such as loose clothing, beyond the rigid constraints of template-based approaches.

\end{itemize}

\section{Related Work}
\label{sec:related_work}

\noindent\textbf{3D Avatar Reconstruction.} High-fidelity 3D avatar reconstruction has traditionally relied on multi-view calibrated systems to capture subjects in a canonical space~\cite{liu2021neural, bagautdinov2021driving, remelli2022drivable, peng2021neural, habermann2021, peng2021animatable, xu2022sanerf, HVTR:3DV2022, 2021narf, li2022tava, zhang2021stnerf, ARAH:ECCV:2022, chen2024meshavatar, saito2024rgca, zheng2023avatarrex, li2023posevocab, 10.1145/3697140, shen2023xavatar, chen2025taoavatar}. Early efforts utilized neural implicit representations such as NeRF~\cite{mildenhall2020nerf, yariv2021volume} to model geometry and appearance, but required time-consuming per-avatar optimization~\cite{peng2021animatable, xu2022sanerf, zhang2021stnerf, ARAH:ECCV:2022}. Recently, the integration of 3D Gaussian Splatting (3DGS)~\cite{kerbl3Dgaussians} has significantly accelerated this process, enabling real-time rendering and faster convergence for person-specific avatars~\cite{li2024animatablegaussians, zielonka2023drivable3dgaussianavatars, moreau2024human, Pang_2024_CVPR, jung2023deformable3dgaussiansplatting}. While these methods can achieve high fidelity by integrating body priors~\cite{smplx:2019, weng2022humannerf, guo2023vid2avatar, moon2024exavatar}, they remain inherently subject-specific and fail to generalize to unseen identities. 

\vspace{1mm}\noindent
To overcome these limitations, feed-forward methods are designed to reconstruct avatars instantly. For static reconstruction, early regression-based efforts utilized pixel-aligned features to infer 3D surfaces or Gaussians from a single view~\cite{saito2019pifu, saito2020pifuhd, xiu2022icon, alldieck2022phorhum, xiu2023econ, huang2020arch, he2021arch++, zheng2021pamir}. However, they often suffer from texture artifacts and self-occlusions. Another line of research leverages sparse-view~\cite{tao2021function4d, zheng2023learning, zheng2024gpsgaussian, kwon2024ghg, Zhao_2022_CVPR, Chen_2023_CVPR, chen2022gpnerf} or novel-view synthesis from generative models~\cite{ho2024sith, li2024pshuman, chen2025synchuman} to achieve high-fidelity results. While impressive, these static reconstructions often lack dynamic motion priors, leading to unrealistic artifacts during animation. Driven by large-scale datasets, recent animatable feed-forward works~\cite{zhuang2024idol, qiu2025lhm, guo2023vid2avatar,li2026largescalecodecavatarsunreasonable} leverage scalable transformers to encode subjects into structured latent spaces, showing promising generalization capability. Our work builds upon these generalizable backbones but significantly extends their capability from static reconstruction to LBS-free, universal implicit animation.

\vspace{2mm}
\noindent\textbf{3D Avatar Animation.}
Conventional 3D aniamtion predominantly rely on LBS tied to parametric models like SMPL~\cite{loper2015smpl, SMPL-X:2019}. To model non-rigid dynamics such as loose garments, recent works augment the LBS prior with pose-dependent Gaussian mapping or neural deformation fields~\cite{li2024animatablegaussians, li2024gaussianbodyclothedhumanreconstruction, chen2025taoavatar, qiu2025lhm}. AniGS~\cite{qiu2024AniGS} further studies animatable Gaussian avatars from a single image, but still relies on a human-prior-driven animation pipeline. However, this tight coupling with the SMPL topology makes these methods highly sensitive to pose estimation errors and restricts them to standard humanoid structures. Although some attempts, such as TAVA~\cite{li2022tava} and GART~\cite{lei2023gart}, relax these constraints via skeleton-based deformations, they still necessitate person-specific multi-view fitting. Recent feed-forward reconstruction models such as PF-LHM~\cite{qiu2025pflhm} improve pose-free avatar reconstruction from casually captured images. Most recently, HumanRAM~\cite{yu2025humanram} introduced an implicit animator for feed-forward 2D reposing, but it lacks a cohesive 3D representation and remains bottlenecked by human template priors. 

\vspace{2mm}
\noindent\textbf{2D Human Animation.} 2D Human animation aims to generate novel-pose images from one or more reference images. Early works~\cite{chan2019everybody, liu2019liquid, zhao2022thin} primarily formulated this task as signal-driven image-to-image translation. Recently, diffusion-based methods~\cite{hu2024animate, zhang2024mimicmotion, zhu2024champ, xu2024magicanimate, wang2024disco, shao2024human4dit} have achieved remarkable success by leveraging the powerful generative priors of 2D diffusion models. While these methods exhibit impressive versatility in handling diverse driving signals (e.g., poses, depth, or edges), they are inherently restricted on 2D image plane, often suffering from temporal flickering, view inconsistency, and a lack of true 3D geometric control, while the iterative denoising process remains computationally expensive. 

\vspace{1mm}\noindent
In contrast to the existing 2D or 3D works, we are the first to bridge the gap between the versatility of 2D driving signal and the structural integrity of 3D modeling. Our proposed LUNA is a fully LBS-free framework that completely discards template dependency and efficiently maps 2D driving signals to 3D avatar motion. 
Our method not only captures complex non-rigid dynamics with superior flexibility but also achieves robust cross-modality generalization that previous 3D animators could not reach.

\section{Method}
\label{sec:method}

\begin{figure}[ht]
    \centering    \includegraphics[width=\linewidth]{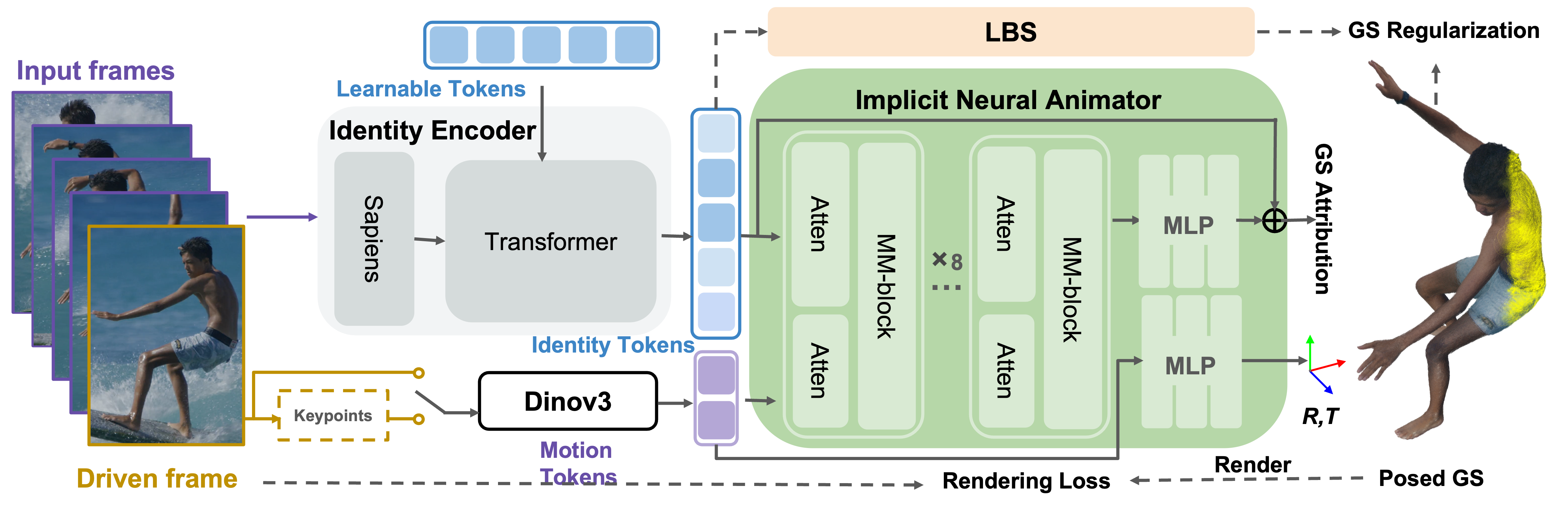}
    \caption{\textbf{Overview}. Given $N$ unposed multi-view identity images and a 2D driving signal, \methodname first reconstructs canonical 3D Gaussians with an \textit{Identity Encoder}. A transformer-based \textit{Implicit Neural Animator} then maps them to posed space conditioned on the driving signal. During training, the driving image is randomly sampled across modalities (RGB, keypoints or sketches). }
    \label{fig:pipeline}
\end{figure}

Given a sparse set of unposed identity images $I_{id} = \{I_{i}\}_{i=1}^N$ and a single 2D driving image $I_d$, our goal is to synthesize a high-fidelity 3D human avatar that can be animated by $I_d$ end-to-end manner. As shown in Fig.~\ref{fig:pipeline}, \methodname has two components: an \textit{Identity Encoder} (Sec.~\ref{sec:canonical_rep}) that lifts the input into a canonical set of 3D Gaussian, and an \textit{Implicit Neural Animator} (Sec.~\ref{subsec:animator}) that predicts disentangled deformations and attribute residuals to pose the avatar. The model is trained with a hybrid supervision strategy (Sec.~\ref{subsec:loss}). Although $I_{id}$  and $I_d$ share the same identity during training, our modelgeneralizes to cross-identity driving at inference. 

\subsection{Canonical Identity Encoder}
\label{sec:canonical_rep}
Inspired by feed-forward human reconstruction model~\cite{qiu2025lhm}, we design an Identity Encoder that lifts the reference images $I_{id}$ into canonical 3D Gaussians, together with semantic features that  condition the subsequent driving.

\vspace{2mm}
\noindent \textbf{Semantic Query Tokens.} To inject human-centric structural priors, we anchor our canonical avatar to a template topology. Specifically, we define $K$ learnable query tokens, each associated with a vertex of the template mesh. To encode anatomical semantics, each token is augmented with a learnable positional embedding $Q \in \mathbb{R}^{K \times C}$ and a semantic label embedding $E_{sem} \in \mathbb{R}^{K \times C}$. The resulting 3D query tokens are formulated as $T_{gs} = Q + E_{sem}$, where $C$ denotes the feature dimension. This design encourages a consistent correspondence between semantic parts and Gaussian primitives, improving identity stability under large pose and viewpoint changes.

\vspace{2mm}
\noindent \textbf{Image Tokenization.} In parallel, we extract fine-grained identity features from the multi-view inputs $I_{id}$. We use a pre-trained Sapiens~\cite{khirodkar2024sapiens} encoder to tokenize each view into patch features, and then aggregate them across views to obtain identity image tokens $T_{img} \in \mathbb{R}^{M \times C}$, where $M$ is the number of image tokens. Following~\cite{qiu2025lhm}, we include both body tokens and face tokens in  $T_{img}$. 

\vspace{2mm}
\noindent \textbf{Joint Feature Fusion and Canonical Decoding.} 
To lift the 2D observations into a unified 3D canonical representation, we fuse the 3D query tokens $T_{gs}$ with the image tokens $T_{img}$ using Multimodal Transformer (MM-Transformer)~\cite{esser2024scaling} blocks. Following~\cite{qiu2025lhm}, each block contains intra-modal self-attention and cross-modal attention to enable cross-modal feature exchange. After $L$ blocks, we obtain canonical identity tokens $T_{can}$, which are decoded by a lightweight MLP into canonical Gaussians ${G}_{can}$. Each Gaussian primitive is parameterized as:
\begin{equation}
g = \{\mu, q, s, \alpha, c\}
\end{equation}
where $\mu \in \mathbb{R}^3$ is the center, $q \in \mathbb{R}^4$ is the rotation quaternion, $s \in \mathbb{R}^3$ is the scale, $\alpha \in \mathbb{R}$ is the opacity, and $c$ is the color. Gaussians $G_{can}$ and tokens $T_{can}$ are then passed to the Implicit Neural Animator.

\subsection{Implicit Neural Animator}
\label{subsec:animator}

The Implicit Neural Animator maps the canonical avatar to posed space conditioned on a 2D driving signal $I_d$. To stabilize learning and improve expressiveness, \methodname disentangles motion into a global rigid transformation and local, per-Gaussian deformations, both predicted in a purely feed-forward manner.

\vspace{2mm}
\noindent \textbf{Motion Tokenization.} We extract motion cues from $I_d$ using a pre-trained DINOv3~\cite{simeoni2025dinov3} encoder, producing motion tokens $T_{motion}$. We choose DINOv3 over Sapiens for two reasons: (i) animation primarily depends on kinematic layout rather than fine-grained identity or texture, and (ii) DINOv3 generalizes better to out-of-domain, non-photorealistic driving signals such as 2D keypoints and abstract sketches.

\vspace{2mm}
\noindent \textbf{Global Transformation Prediction.} We model the avatar’s coarse spatial alignment with a global rotation $R$ and translation $T$. We first aggregate the DINOv3 motion tokens into a global descriptor and feed it to two lightweight MLP heads. Directly regressing Euler angles is unstable due to discontinuities at $\pm\pi$, so we predict rotation in a continuous trigonometric form. Specifically, for each axis $i \in \{x,y,z\}$, the rotation head outputs a 2D vector, 
\begin{equation}
(s_i, c_i) = \text{Tanh} (\text{MLP}_{rot}(T_{motion})) 
\end{equation}
which is normalized and then composed into a valid rotation matrix $R \in \mathbb{R}^{3\times3}$. For translation, regressing absolute coordinates can exhibit scale drift. Instead, we predict a normalized offset $\hat{t}$ and recover translation via dataset denormalization:
\begin{equation}    
T = \sigma_T \cdot \hat{t} + \mu_T, \hat{t}=\text{Tanh} (\text{MLP}_{trans}(T_{motion}))
\end{equation}
where $\mu_T$ and $\sigma_T$ are the mean and standard deviation computed from the training data distribution.

\vspace{1mm}
\noindent \textbf{Deformed Avatar Gaussian.} Global $R$ and $T$ provide coarse alignment, while local deformations capture the coupling of articulated motion and non-rigid dynamics. We first project the canonical tokens $T_{can}$ to a lower-dimensional latent and concatenate them with learnable motion queries $Q_{motion} \in \mathbb{R}^{K \times (C/2)}$ to form identity queries:
\begin{equation}
T_{id}=\text{MLP}_{proj}(T_{can}) \parallel Q_{motion}
\end{equation}
where $\parallel$ denotes channel-wise concatenation. We fuse $T_{id}$ with the motion tokens $T_{motion}$ using the MM-Transformer, and decode per-Gaussian residuals with a lightweight MLP: position offset $\Delta\mu$, rotation residual $\Delta q$, and color residual $\Delta c$. For stability, we keep scale $s$ and opacity $\alpha$ fixed during animation. The posed Gaussians are:
\begin{equation}
g_{posed}=\{\mu_p, q_p, s, \alpha, c_p\}
\end{equation}
where $\mu_p = R(\mu+\Delta\mu)+T$, $q_p = q_R \otimes \Delta q \otimes q$, and $c_p = c + \Delta c$. Here $\otimes$ denotes quaternion multiplication and $q_R$ is the quaternion corresponding to the global rotation $R$.

\subsection{Hybrid Supervision Strategy}
\label{subsec:loss}
Our training corpus mixes samples with parametric body-fitting annotations and large-scale unannotated videos. We first train the Identity Encoder on labeled data by extending LHM to a multi-view setting (MV-LHM), and then train the Neural Animator using both labeled and unlabeled data. Concretely, we sample the two data types within each mini-batch and optimize a balanced hybrid objective for stable training.

\vspace{2mm}
\noindent \textbf{Photometric Rendering Loss.} For all samples, we rasterize the posed 3D Gaussians into an image $I_{render}$ and enforce photometric consistency with the driving frame $I_d$:
\begin{equation}
\mathcal{L}_{render} = \mathcal{L}_1+\mathcal{L}_{mask} +  \mathcal{L}_{LPIPS}
\end{equation}
where $\mathcal{L}_1$ is the pixel-wise $L_1$ loss, $\mathcal{L}_{mask}$ is the foreground mask loss, and $\mathcal{L}_{LPIPS}$ is the perceptual loss.

\vspace{2mm}
\noindent \textbf{LBS Distillation Regularization.} For samples with parametric pose labels, we use an LBS-based teacher to provide soft structural targets. The teacher produces pseudo attributes for posed Gaussians, including positions $\hat{\mu}$, rotations $\hat{q}$, and colors $\hat{c}$. We guide the animation transformer with the distillation loss:
\begin{equation}
\mathcal{L}_{distill} = \|\mu_p - \hat{\mu}\|_1 + \lambda_{q} \mathcal{L}_{rot}(q_p, \hat{q}) + \lambda_{c} \|c_p - \hat{c}\|_1
\end{equation}
where $\mathcal{L}_{rot}$ denotes the cosine distance between quaternions.

\vspace{2mm}
\noindent \textbf{Projection Loss for Transformation.} A natural strategy is to supervise the predicted global rotation and translation on the annotated subset and rely on photometric rendering to generalize to unlabeled data. In practice, direct supervision is stable for rotation, so we supervise $R$ via its representation:
\begin{equation}
    \mathcal{L}_{R} = \|(s, c) - (\hat{s}, \hat{c})\|_1
\end{equation}
However, for translation $T$, even marginal estimation errors (especially along the depth axis) can induce severe spatial misalignment, thus dominating the loss gradients and destabilizing the training. We therefore avoid direct 3D supervision on $T$ and instead impose a 2D reprojection loss on posed Gaussian centers:
\begin{equation}
\mathcal{L}_{proj} = \frac{1}{K} \sum_{i=1}^{K} \left\| \Pi(\mu_p^i) - \hat{u}^i \right\|_1    
\end{equation}
where $\Pi(\cdot)$ represents the camera projection.

\vspace{2mm}
\noindent \textbf{Balanced Hybrid Optimization.} 
When mixing labeled and unlabeled data, photometric gradients from the unlabeled majority can overwhelm the structural signal, leading to geometric collapse (see Fig.~\ref{fig:abla_distill_mvfinetuning}). To maintain gradient balance, we form mini-batches with a fixed labeled-to-unlabeled ratio (e.g., $1{:}5$) and upweight the distillation term:
\begin{equation}
    \mathcal{L}_{total} = \mathcal{L}_{render} + \mathcal{L}_{R} + \mathcal{L}_{proj} + \lambda_{distill} \cdot \mathcal{L}_{distill}
\end{equation}
We set $\lambda_{distill}$ inversely proportional to the labeled fraction (e.g., $\lambda_{distill}=5$ for $1{:}5$), preventing the structural gradients from being diluted. This simple reweighting enables learning fine non-rigid details from large-scale unlabeled videos while retaining LBS-anchored geometric stability.

\section{Experiments}
\label{sec:experiments}

\subsection{Implementation Details}

\noindent \textbf{Data.} 
Our training corpus integrates four distinct data sources to ensure robust generalization across diverse identities and poses.
\begin{itemize}
\item \textbf{Video35K}: 35,000 in-the-wild full-body video clips, uniformly sampling 60 frames per clip. Video35K was not collected from a third-party data provider under a licensing agreement that permits machine learning research and publication, and is curated from the same source as the Sapiens pre-training data~\cite{khirodkar2024sapiens}. 
\vspace{1mm}

\item \textbf{iPhone1K}: 1,100 monocular videos captured by iPhone, featuring subjects performing basic turn-around motions. It is a proprietary dataset collected onsite by our team and collaborators, following the phone-scan capture setting used in URAvatar~\cite{li2024uravatar}. All participants provided written informed consent covering the use of their data for machine learning research and publication. We select 1,000 identities and uniformly sample 32 frames per identity for training.
\vspace{1mm}

\item \textbf{Cloth10K}: To explicitly model complex non-rigid dynamics, we curate 10,000 identities with complex garments and reserve 100 distinct clips for evaluation. Cloth10K is a derivative dataset constructed by us from Video35K using a conditional video generation model, and is used under the same applicable licensing terms.
\vspace{1mm}
\item \textbf{Dome}: A multi-view dataset of 900 identities captured in a 3D studio with 200 synchronized cameras. It is a proprietary onsite capture dataset collected by our team and collaborators, and follows the same type of studio capture source as Ava-256~\cite{martinez2024codec}. All participants provided written informed consent covering the use of their data for machine learning research and publication. We uniformly sample 200 frames per identity from the original $\sim$2,000-frame casual motion sequences for training.
\end{itemize}
In total, the training dataset includes over 36,000 identities and 3.7M frames.

\vspace{2mm}
\noindent \textbf{Evaluation.} We evaluate self-reenactment and animation quality on the 100 Cloth10K clips and 100 iPhone sequences. We also benchmark on the widely used NeuMan~\cite{jiang2022neuman} dataset, which was obtained from its official public release and used following the protocol described by its authors. For optimization-based baselines (Vid2Avatar~\cite{guo2023vid2avatar}, ExAvatar~\cite{moon2024exavatar}), we strictly follow the official frame splits for training. For feed-forward methods, we uniformly select one or four views from the training splits as reference input. All methods are evaluated on the official test splits. We further quantitatively evaluate cross-identity driving on NeuMan.

\begin{figure}[t]
    \centering    \includegraphics[width=\linewidth]{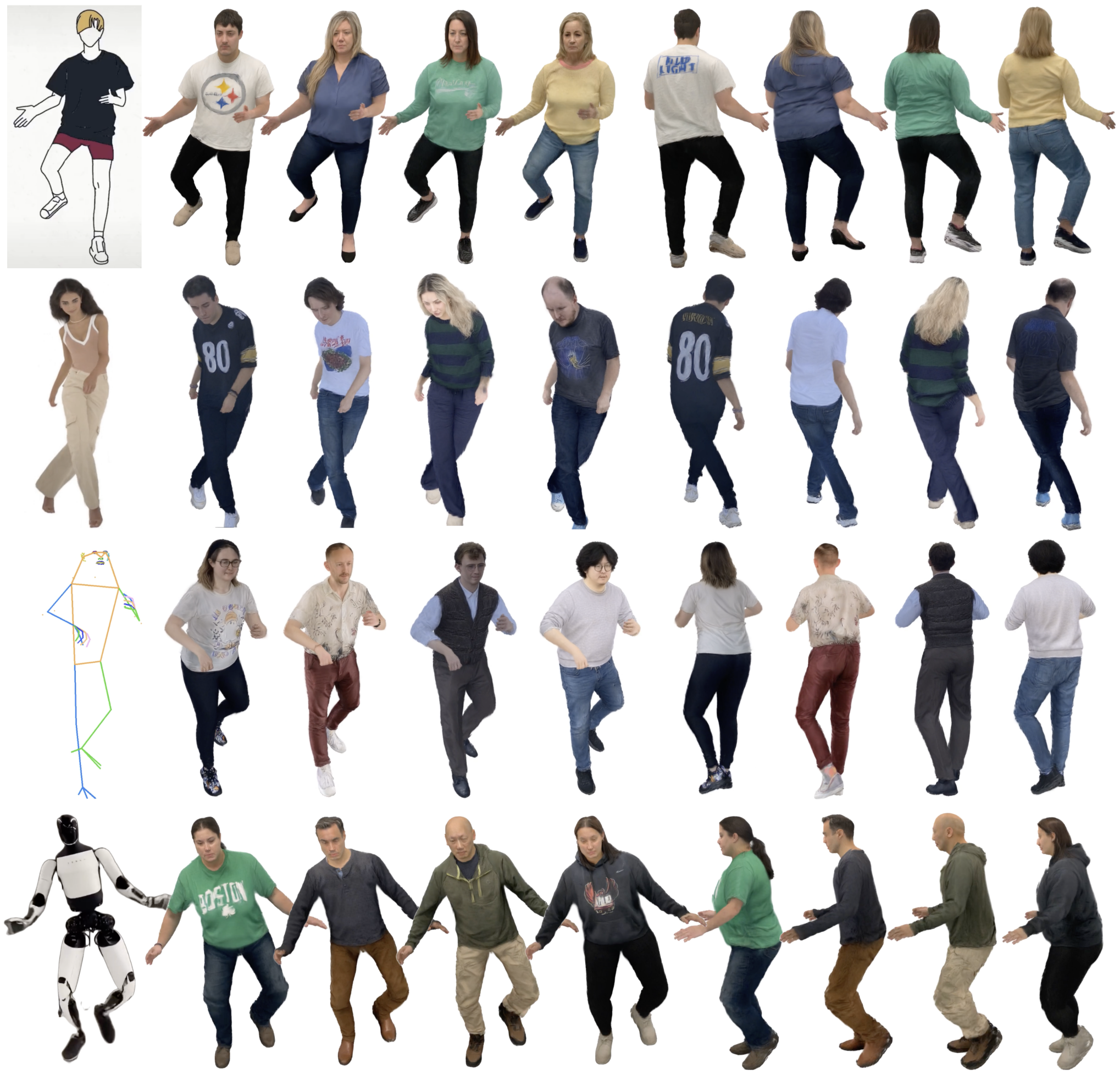}
    \footnotesize\leftline{~~~~~~Driver~\qquad\qquad\qquad\qquad\qquad\qquad~~~Posed 3D Human~}
    \caption{\textbf{Our results with diverse 2D driving signals}, including sketch image, color image, 2D keypoint and other out-of-domain images.}
    \label{fig:our_results}
\end{figure}

\begin{figure}[t]
    \centering    
    \includegraphics[width=\linewidth,height=0.8\linewidth]{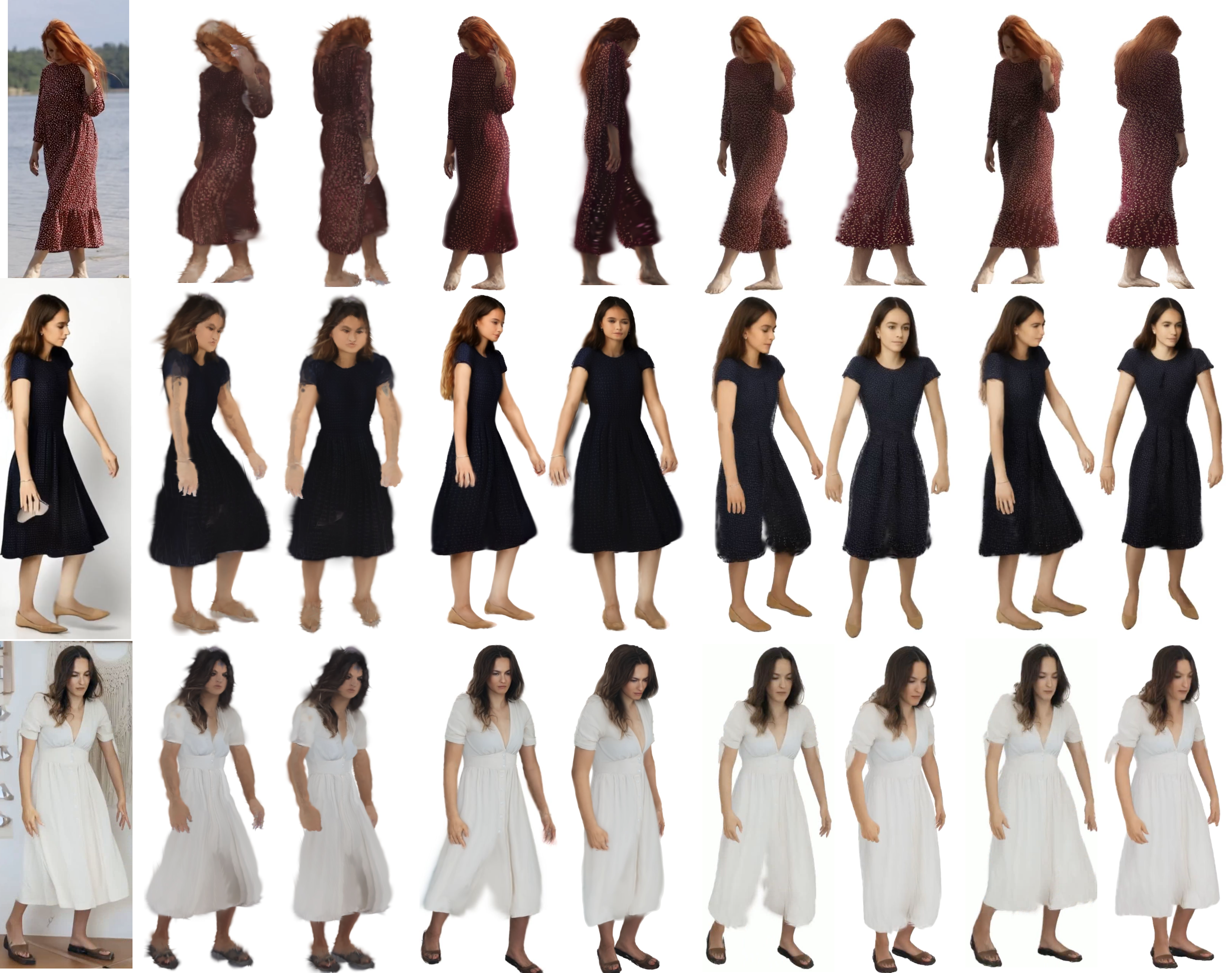}
    \footnotesize\leftline{Reference~\qquad~~IDOL~\cite{zhuang2024idol}~\qquad~~LHM~\cite{qiu2025lhm}~~\qquad\qquad~~MV-LHM~~~\qquad\quad~~~~Ours~~}
    \caption{\textbf{Qualitative comparisons on Cloth10K.} LBS-dependent approach (IDOL and LHM) suffer from severe ID shift ot structural tearing. In contrast, \methodname preserves the structural integrity and continuous topology of the fabric.}
    \label{fig:loose_cloth}
\end{figure}

\vspace{2mm}
\noindent \textbf{Baselines and Metrics.} We compare against monocular optimization methods (Vid2Avatar~\cite{guo2023vid2avatar}, ExAvatar~\cite{moon2024exavatar}) and recent feed-forward models (IDOL~\cite{zhuang2024idol}, LHM~\cite{qiu2025lhm}, UP2YOU~\cite{cai2025up2you}). We also include our extended MV-LHM to benchmark multi-view inputs. Photometric quality is measured via PSNR, L1 error, and LPIPS. To quantify motion stability and temporal plausibility, we employ Mean Acceleration Error~(MAE)~\cite{Kanazawa_2019_CVPR} and Mean Squared Jerk (MSJ)~\cite{du2023agrol}. MAE evaluates the second-order temporal derivative of 3D Gaussian trajectories, reflecting overall motion stability:

\begin{equation}
E_{acc} = \frac{1}{N (T - 2)} \sum_{i=1}^{N} \sum_{t=2}^{T-1} \left\| \frac{\mathbf{x}_{i,t+1} - 2\mathbf{x}_{i,t} + \mathbf{x}_{i,t-1}}{\Delta t^2} \right\|    
\end{equation}
where $\mathbf{x}_{i,t}$ denotes the 3D position of the $i$-th Gaussian at frame $t$, and $\Delta t$ is the time interval. MSJ measures the third-order derivative, serving as a highly sensitive indicator for high-frequency "jittering" artifacts,

\begin{equation}
E_{jerk} = \frac{1}{N (T - 3)} \sum_{i=1}^{N} \sum_{t=3}^{T-1} \left\| \frac{\mathbf{x}_{i,t+1} - 3\mathbf{x}_{i,t} + 3\mathbf{x}_{i,t-1} - \mathbf{x}_{i,t-2}}{\Delta t^3} \right\|^2    
\end{equation}
Lower values for both metrics indicate smoother, physically coherent animations.

\vspace{2mm}
\noindent \textbf{Hyperparameters.} The Identity Encoder is initialized from our MV-LHM, pre-trained on $\sim$1M pose-annotated videos (details in Appendix). The Neural Animator is trained in two progressive stages. 
At Stage 1 (monocular pre-training), the model is trained on Video34K, iPhone1K, and the Cloth10K training split using 64 A100 GPUs (batch size 64) for 30K iterations with a learning rate of $4 \times 10^{-4}$. The training begins with a 1,000-iteration warmup supervised only by the global rotation and projection loss ($\mathcal{L}_R + \mathcal{L}_{proj}$), after which the full objective $\mathcal{L}_{total}$ is applied.  At stage 2, we fine-tune the model on the Dome dataset using 16 A100 GPUs for 30K iterations with a decayed learning rate of $1 \times 10^{-4}$. 
The hyper-parameters are set to $\lambda_q=\lambda_c=0.5$, $K=8192$, $C=1024$, $N=4$. We employ MHR~\cite{MHR:2025} as body template in our experiments.

\begin{table}[t]
\centering
\caption{\textbf{Quantitative comparisons of reconstruction quality.} We evaluate \methodname on our collected Cloth10K~(focusing on non-rigid garments) and the public NeuMan~\cite{jiang2022neuman} dataset. Bold indicates the best performance, and $\ast$ denotes our implementation for multiview inputs.}
\label{tab:main_comparison}
\begin{tabular}{c l ccc ccc}
\toprule
\multirow{2}{*}{\textbf{Type}} & \multirow{2}{*}{\textbf{Method}} & \multicolumn{3}{c}{\textbf{Cloth10K~ (Non-rigid)}} & \multicolumn{3}{c}{\textbf{NeuMan }} \\
\cmidrule(lr){3-5} \cmidrule(lr){6-8}
& & PSNR$\uparrow$ & L1$\downarrow$ & LPIPS$\downarrow$ & PSNR$\uparrow$ & L1$\downarrow$ & LPIPS$\downarrow$ \\
\midrule
\multirow{2}{*}{Optim. based} 
& Vid2Avatar~\cite{guo2023vid2avatar}  &18.971. &0.042 &0.157 &26.853 & 0.012 & 0.017 \\
& ExAvatar~\cite{moon2024exavatar}  &19.533. &0.041 &0.153 &\textbf{31.270} & \textbf{0.009} & \textbf{0.009}    \\
\midrule
\multirow{5}{*}{Feed-forward} 
& IDOL~\cite{zhuang2024idol}     &17.430 &0.067 &0.213 & 24.700 & 0.037 & 0.051             \\
& LHM~\cite{qiu2025lhm}  &19.201 &0.047 &0.167  & 25.310 & 0.029 & 0.039           \\
& UP2YOU~\cite{cai2025up2you}      &19.745. &0.047 &0.149 & 25.492 & 0.026 & 0.041    \\
& MV-LHM$^\ast$                         & 20.124 & 0.038 & 0.158 & 26.832 & 0.017 & 0.023\\
\cmidrule{2-8}
& \textbf{\methodname (Ours)}                 & \textbf{22.072} & \textbf{0.027} & \textbf{0.131} & 26.819 & 0.015 & 0.023 \\
\bottomrule
\end{tabular}
\end{table}

\begin{table}[t]
\centering
\caption{\textbf{Quantitative evaluation of cross-identity driving on NeuMan.} For Cross-ID, we animate 10 curated in-the-wild identity images with NeuMan motions and use the generated videos as driving signals. The right subtable reports 2D keypoint error extracted by Sapiens and normalized by the person bounding box.}
\label{tab:cross_id_keypoint}
\begin{minipage}{0.48\linewidth}
\centering
\footnotesize
\begin{tabular}{lccc}
\toprule
Method & PSNR$\uparrow$ & L1$\downarrow$ & LPIPS$\downarrow$ \\
\midrule
LHM++~\cite{qiu2025lhm} & 25.744 & 0.023 & 0.032 \\
\methodname (Cr-ID) & 26.800 & 0.017 & 0.027 \\
\methodname (Self) & \textbf{26.819} & \textbf{0.015} & \textbf{0.023} \\
\bottomrule
\end{tabular}
\end{minipage}
\hfill
\begin{minipage}{0.45\linewidth}
\centering
\footnotesize
\begin{tabular}{lcccc}
\toprule
Method & IDOL & LHM & MV-LHM & Ours \\
\midrule
Err.($\%$)$\downarrow$ & 28.7 & 26.7 & 26.7 & \textbf{17.3} \\
\bottomrule
\end{tabular}
\end{minipage}
\end{table}
\subsection{Results}
\label{sec:comparisons}


\textbf{Reconstruction and Animation.}
By directly extracting motion cues from 2D signals, \methodname bypasses explicit 3D pose estimation and rigid SMPL fitting. This end-to-end paradigm inherently avoids the error accumulation and domain shift vulnerabilities of LBS based animation. Consequently, our approach enables flexible 3D animation driven by highly diverse, cross-modality sources, including hand-drawn sketches, 2D skeletons, and out-of-domain humanoid images (Fig.~\ref{fig:teaser}, \ref{fig:our_results}). We refer readers to the supplementary video for dynamic visualizations.

\vspace{1mm}
\noindent\textbf{Cross-identity driving and pose accuracy.}
Although \methodname is trained with matched source and driving identities, it generalizes to cross-identity driving without any explicit transfer module. To evaluate this setting on NeuMan, we curate 10 in-the-wild human images for each video and use Wan-Animate~\cite{cheng2025wananimate} to animate them with the corresponding NeuMan motions. These generated videos are then used as driving signals for \methodname. As shown in Tab.~\ref{tab:cross_id_keypoint}, cross-identity driving remains close to self-reenactment and outperforms LHM++~\cite{qiu2025lhm}, indicating that the animator primarily extracts motion rather than appearance from the driving signal. We also evaluate normalized 2D keypoint error using Sapiens detections; \methodname achieves the lowest error, further confirming that direct implicit driving improves motion correctness.


\vspace{3mm}
\noindent\textbf{Comparison of reconstruction quality with LBS-based methods.}
We conduct qualitative comparisons between state-of-the-art optimization-based and feed-forward works with ours on 100 Cloth10K samples. As shown in Fig.~\ref{fig:loose_cloth}, LBS-based methods like IDOL and LHM exhibit severe ID shift or catastrophic geometric collapse on loose garments due to the skeletal binding constraints. By discarding the rigid LBS, our implicit animator can naturally capture the non-linear sliding of fabric and significantly mitigate garment tearing artifacts. Quantitatively, as reported in Tab.~\ref{tab:main_comparison}, our approach significantly outperforms all baselines across photometric metrics on the Cloth10K dataset, demonstrating a clear advantage in modeling complex motion of non-rigid garments. On the NeuMan dataset, which primarily features standard walking motions and tighter clothing, \methodname achieves comparable performance with feed-forward baselines, proving that our design maintains high fidelity on simpler, highly constrained scenarios as well.

\begin{figure}[t]
    \centering    \includegraphics[width=\linewidth]{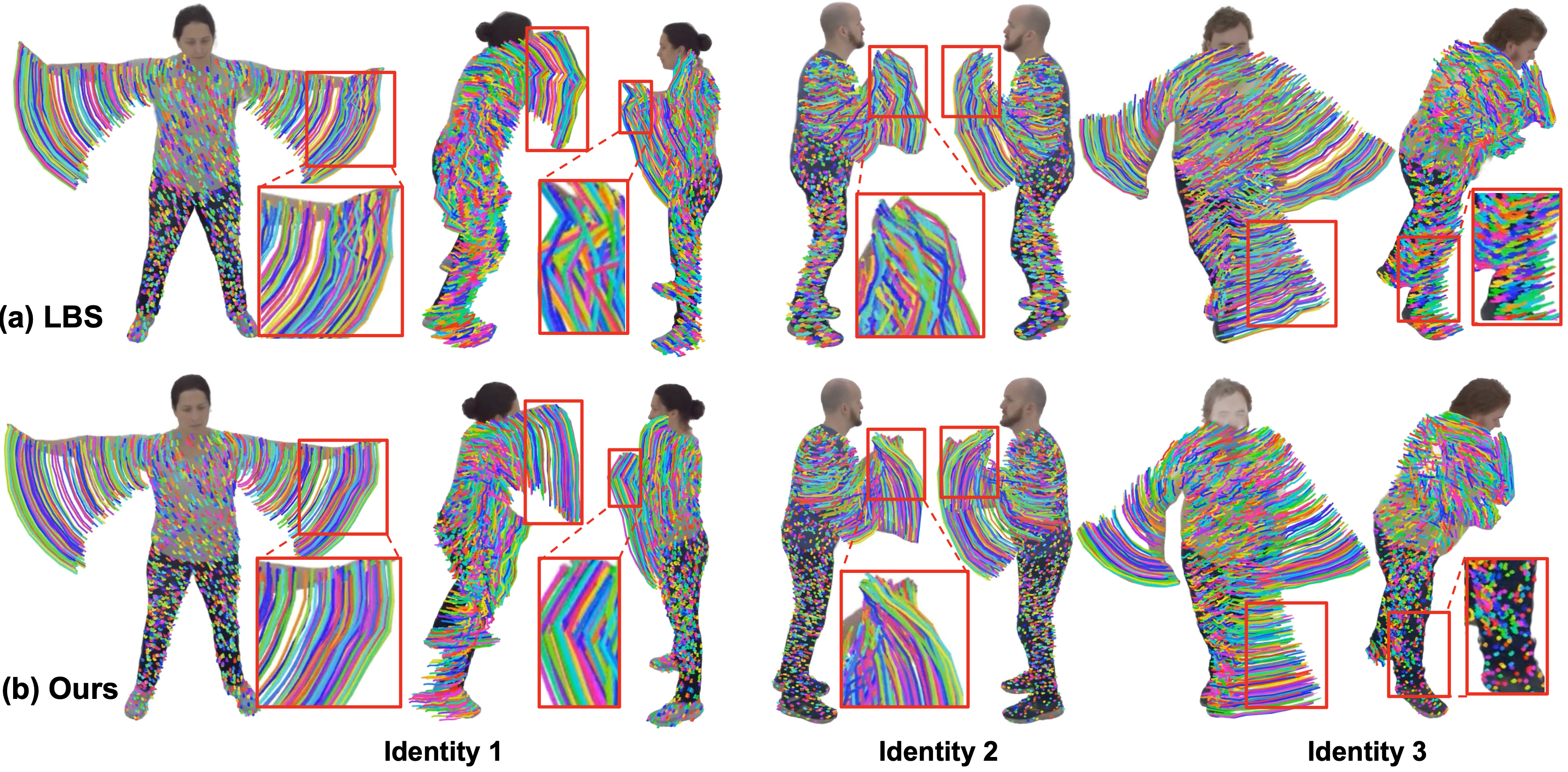}
    \caption{\textbf{Qualitative comparisons of animation smoothness}. We visualize the per-gaussian trajectory of (a) LBS-based animation baseline (b) our neural approach. The LBS-based methods suffer from erratic trajectories due to pose estimation errors and rigid skinning constraints. For all methods, we use SAM-3D-Body~\cite{yang2026sam3dbody} to extract body parameters from driving videos and transform them to SMPL(X) for avatar animation.
    }
    \label{fig:gs_streamline}
\end{figure}

\begin{table}[t]
\centering
\caption{\textbf{Quantitative comparison of motion quality.} We compare our neural animator and various baselines. Due to the error and ambiguity of monocular pose estimation, LBS-based methods exhibit higher motion jitter, whereas our method achieves superior smoothness.}
\label{tab:msj_mae}
\begin{tabular}{lcc|cccc|c}
\toprule
\textbf{Methods}  & IDOL & UP2YOU & Vid2Avatar & ExAvatar  & LHM & MV-LHM$^\ast$ & \textbf{Ours} \\
\midrule
\textbf{Animation} & \multicolumn{2}{c|}{LBS} & \multicolumn{4}{c|}{LBS + Neural} & \textbf{Neural} \\
\midrule
MSJ ($\downarrow$) & 0.0315 & 0.0321 & 0.0237 & 0.0231 & 0.0214 & 0.0144 & \textbf{0.0032} \\
MAE ($\downarrow$) & 0.0591 & 0.0591 & 0.0493 & 0.0532 & 0.0522 & 0.0477 & \textbf{0.0225} \\
\bottomrule
\end{tabular}
\end{table}

\vspace{3mm}
\noindent\textbf{Comparisons of animation quality.} Since frame-wise photometric metrics fail to reflect temporal artifacts, we quantitatively evaluate animation stability using MAE and MSJ. As reported in Tab.~\ref{tab:msj_mae}, \methodname significantly outperforms all pure LBS and hybrid baselines, effectively mitigating the frame-to-frame jitter caused by noisy pose estimation and rigid skinning constraints and achieving a 4.5$\times$ reduction over the strongest baseline (MV-LHM) on MSJ. Notably, the gains are consistent across identities and motions, indicating robust temporal behavior rather than isolated improvements.

\vspace{1mm}
Qualitatively, we visualize the per-Gaussian motion trajectories as streamlines (Fig.~\ref{fig:gs_streamline}). Trajectories produced by LBS-based methods (Fig.~\ref{fig:gs_streamline}a) exhibit severe "zig-zag" artifacts and erratic spatial shifts, particularly around high-dynamic regions. In contrast, our approach (Fig.~\ref{fig:gs_streamline}b) yields smooth, continuous streamlines that better preserve local coherence over time. This advantage is especially evident in the nearly stationary pose (Identity 3, rightmost column), where motion trajectories should ideally degenerate into scatter points. The baseline instead produces spurious, chaotic streamlines induced by pose estimation noise, while \methodname remains near-static and visually stable.

\begin{figure}[b!]
    \centering
    \includegraphics[width=\linewidth]{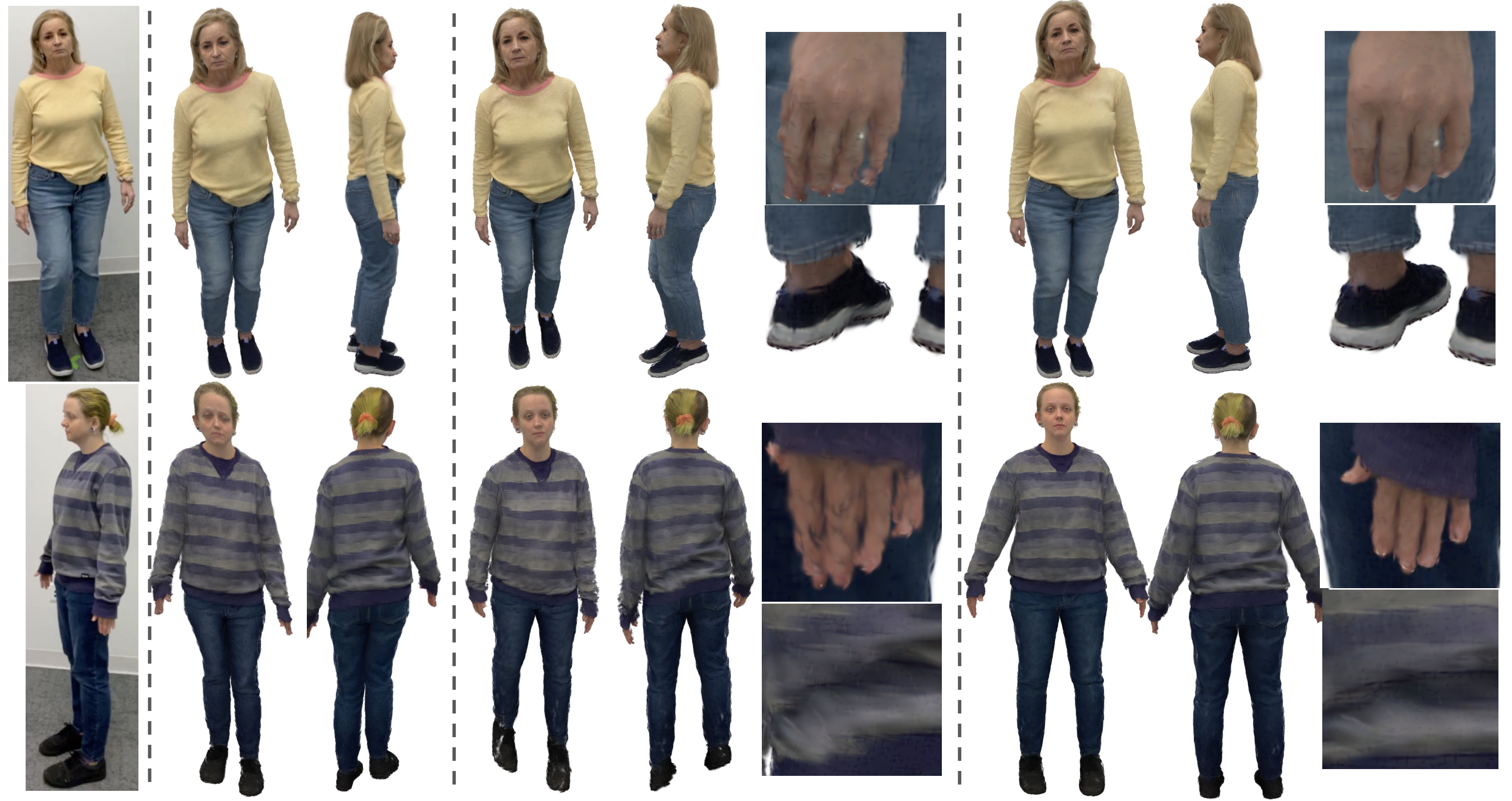}
    \footnotesize\leftline{~Driver~~~w/o distill reg.~~~~~w/o multiview finetuning~~\qquad~~~\qquad~~~~Ours~~}
    \caption{\textbf{Ablations of distillation regularization and multiview finetuning.}}
    \label{fig:abla_distill_mvfinetuning}
\end{figure}
\subsection{Ablation}
\label{sec:ablation}

\smallskip
\noindent\textbf{Effect of distillation regularization.}
As shown in Fig.~\ref{fig:abla_distill_mvfinetuning}, removing LBS-guided distillation (\textit{w/o distill reg.}) causes severe depth collapse and flattening artifacts, degrading both shape coherence and motion consistency. This confirms that while LBS is too rigid for final non-rigid animations, its structural prior is indispensable for resolving early-stage depth ambiguity and establishing global geometric stability.

\noindent\textbf{Effect of multiview finetuning.} Monocular supervision (\textit{w/o multiview finetuning}) inherently suffers from perspective distortions and tends to produce floating Gaussian splats, leading to implausible renderings (Fig.~\ref{fig:abla_distill_mvfinetuning}). Incorporating multi-view finetuning directly eliminates these perspective ambiguities and enforces the implicitly animated 3D Gaussians to be more spatially compact, significantly elevating the overall visual quality.

\begin{figure}[t]
  \centering
  \begin{minipage}[t]{0.48\textwidth}
    \centering
    \includegraphics[width=0.8\linewidth]{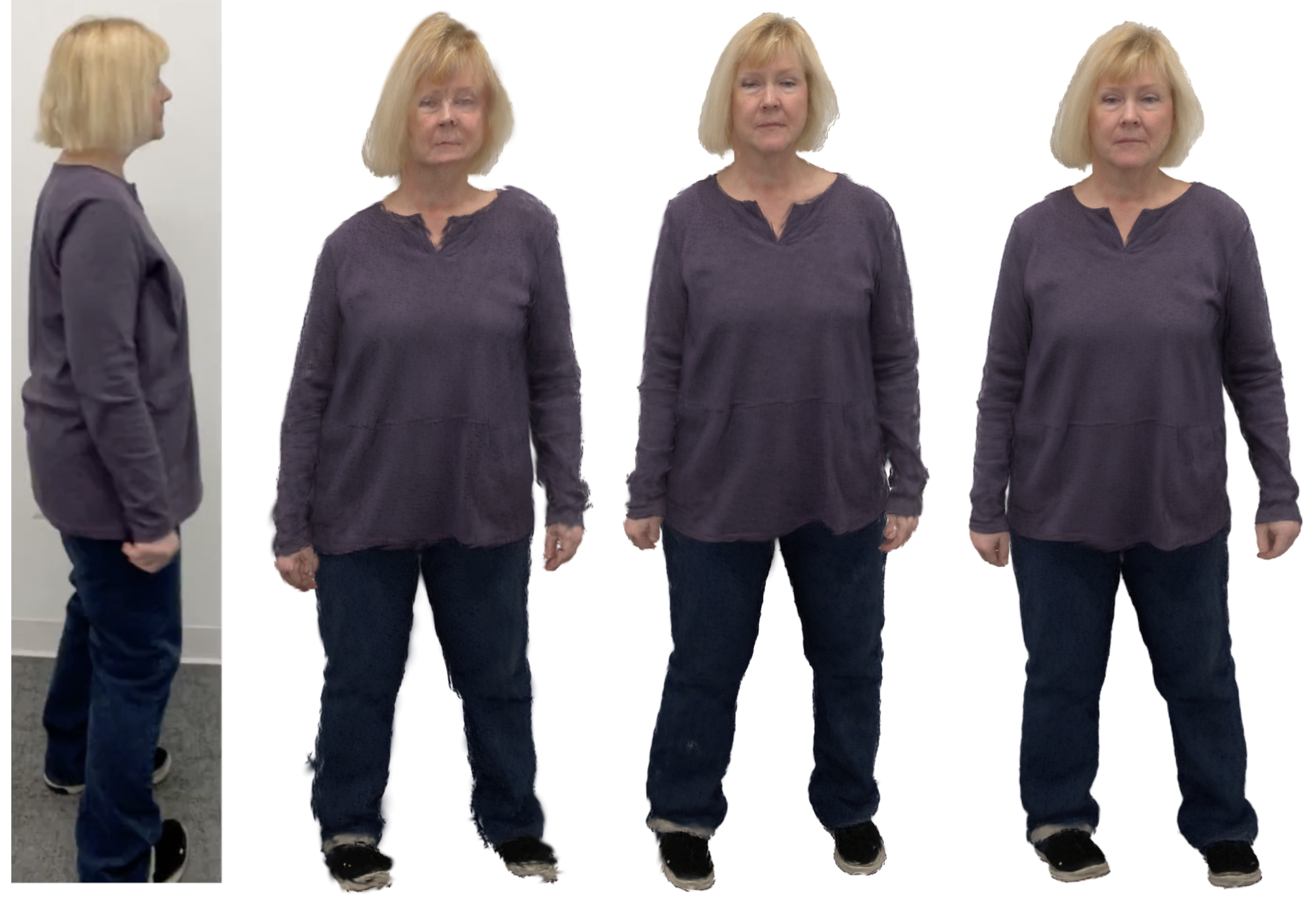}
    \footnotesize\leftline{Driver~\quad~~w/o Rot~~\quad~Ours~\qquad~LBS}
    \caption{\textbf{Ablation of decoupled global rotation prediction.} Without global rotation, the deformation model fails to handle large motion~(e.g. turning round).}
    \label{fig:abla_rot}
  \end{minipage}\hfill
  \begin{minipage}[t]{0.5\textwidth}
    \centering
    \includegraphics[width=\linewidth]{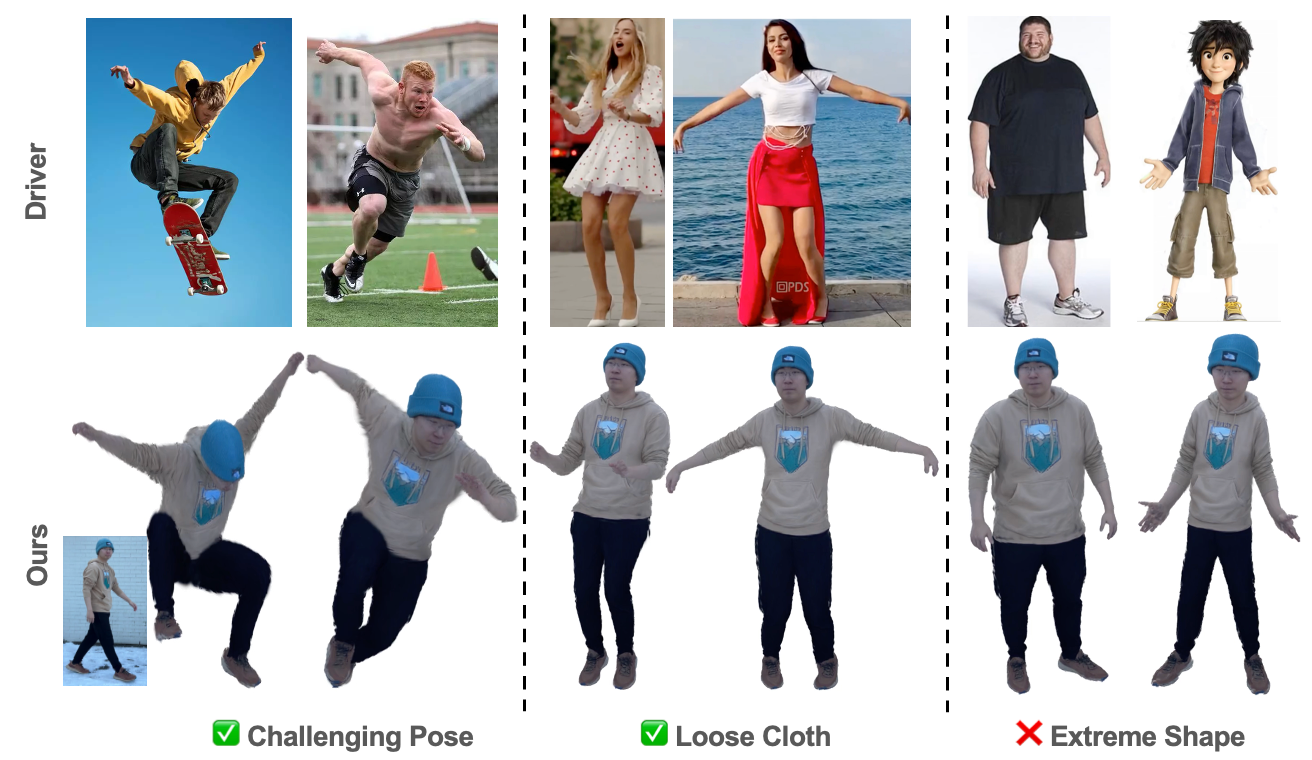}
    \caption{\textbf{Challenging pose-driven cases.} \methodname handles challenging poses and loose-cloth mismatch in many cases, while more extreme pose and body-shape mismatch remain difficult.}
    \label{fig:driver_source_mismatch}
  \end{minipage}
\end{figure}

\vspace{2mm}
\noindent\textbf{Effect of global rotation decoupling.}
We further investigate the design of the decoupled deformation field by training a variant that regresses all Gaussian offsets in a single unified model. As shown in Fig.~\ref{fig:abla_rot}, this setting (\textit{w/o Rot}) forces the model to compensate for large articulations, leading to severe drifting artifacts and ragged boundaries. In contrast, our model focuses exclusively on local non-rigid dynamics, producing sharp results that closely align with LBS reference, as further corroborated by the quantitative results reported in Tab.~\ref{tab:abla_metric}.

\begin{table}[t]
\centering
\caption{\textbf{Ablation study of each components in \methodname training}. We evaluate the contribution of distillation regularization, multiview finetuning, and decoupling rotation from gaussian deformations on 100 samples of iPhone1K and 20 dome identities.}
\begin{tabular}{lccc|ccc}
\toprule
\multirow{2}{*}{\textbf{Setting}} & \multicolumn{3}{c|}{\textbf{iPhone data}} & \multicolumn{3}{c}{\textbf{dome data}} \\
& PSNR $\uparrow$ & L1 $\downarrow$  & LPIPS $\downarrow$ & PSNR $\uparrow$ & L1 $\downarrow$  & LPIPS $\downarrow$ \\
\midrule
w/o Structural Distillation & 21.185 & 0.032 & 0.145 & 21.712 & 0.029 & 0.138 \\
w/o Global Rotation & 23.102 & 0.025 & 0.118 & 23.294 & 0.022 & 0.115 \\
w/o Multiview Finetuning &  23.931& 0.021 & 0.102 & 23.864 & 0.019 & 0.099 \\
\midrule
\textbf{Full Model (Ours)} & \textbf{24.136} & \textbf{0.019} & \textbf{0.094} & \textbf{24.374} & \textbf{0.017} & \textbf{0.093} \\
\bottomrule
\end{tabular}
\label{tab:abla_metric}
\end{table}


\section{Conclusion}
We present \methodname, a novel LBS-free framework for high-fidelity 3D human animation. Bypassing explicit pose estimation entirely, \methodname leverages a Transformer-based animator to directly extract motion cues from diverse 2D signals. To prevent the geometric collapse inherent to LBS-free approaches, we introduce a hybrid supervision strategy utilizing LBS-guided distillation. Finally, the decoupled global-local formulation empowers \methodname to capture complex, unconstrained non-rigid dynamics. Extensive experiments demonstrate \methodname's robust zero-shot generalization across heterogeneous identities, loose clothing, and cross-modality driving sources, establishing a highly flexible paradigm for 3D digital human animation.


\vspace{3mm}
\noindent\textbf{Limitations and Future Work.} 
While LUNA is highly flexible, several directions remain to further strengthen its robustness. First, severe occlusions in the driving video can degrade the extracted 2D semantic cues and occasionally introduce temporal instability. Second, although \methodname handles many challenging pose-driven and loose-cloth cases (Fig.~\ref{fig:driver_source_mismatch}), very extreme poses or large body-shape mismatch can still degrade cross-identity animation because motion and shape are not explicitly disentangled. Moving forward, incorporating explicit temporal sequence modeling, improving occlusion-aware cue extraction, and scaling training with more diverse motions, apparel, and viewpoints are promising avenues to address these cases.


\clearpage 

 
%
%
\bibliographystyle{splncs04}
\bibliography{main}

@String(CVPR  = {IEEE Conf. Comput. Vis. Pattern Recog.})

@String(ECCV  = {Eur. Conf. Comput. Vis.})

@String(NeurIPS = {Adv. Neural Inform. Process. Syst.})

@String(ICML  = {Int. Conf. Mach. Learn.})

@String(TOG   = {ACM Trans. Graph.})

@String(CVPR  = {CVPR})

@String(ECCV  = {ECCV})

@String(NeurIPS = {NeurIPS})

@String(ICML  = {ICML})

@String(TOG   = {ACM TOG})

@inproceedings{saito2019pifu,
  title={Pifu: Pixel-aligned implicit function for high-resolution clothed human digitization},
  author={Saito, Shunsuke and Huang, Zeng and Natsume, Ryota and Morishima, Shigeo and Kanazawa, Angjoo and Li, Hao},
  booktitle={Proceedings of the IEEE/CVF international conference on computer vision},
  pages={2304--2314},
  year={2019}
}

@inproceedings{saito2020pifuhd,
  title={Pifuhd: Multi-level pixel-aligned implicit function for high-resolution 3d human digitization},
  author={Saito, Shunsuke and Simon, Tomas and Saragih, Jason and Joo, Hanbyul},
  booktitle={Proceedings of the IEEE/CVF conference on computer vision and pattern recognition},
  pages={84--93},
  year={2020}
}

@inproceedings{xiu2022icon,
  title={Icon: Implicit clothed humans obtained from normals},
  author={Xiu, Yuliang and Yang, Jinlong and Tzionas, Dimitrios and Black, Michael J},
  booktitle={2022 IEEE/CVF Conference on Computer Vision and Pattern Recognition (CVPR)},
  pages={13286--13296},
  year={2022},
  organization={IEEE}
}

@inproceedings{xiu2023econ,
  title={Econ: Explicit clothed humans optimized via normal integration},
  author={Xiu, Yuliang and Yang, Jinlong and Cao, Xu and Tzionas, Dimitrios and Black, Michael J},
  booktitle={Proceedings of the IEEE/CVF conference on computer vision and pattern recognition},
  pages={512--523},
  year={2023}
}

@inproceedings{ho2024sith,
  title={Sith: Single-view textured human reconstruction with image-conditioned diffusion},
  author={Ho, I and Song, Jie and Hilliges, Otmar and others},
  booktitle={Proceedings of the IEEE/CVF Conference on Computer Vision and Pattern Recognition},
  pages={538--549},
  year={2024}
}

@article{hong2023lrm,
  title={Lrm: Large reconstruction model for single image to 3d},
  author={Hong, Yicong and Zhang, Kai and Gu, Jiuxiang and Bi, Sai and Zhou, Yang and Liu, Difan and Liu, Feng and Sunkavalli, Kalyan and Bui, Trung and Tan, Hao},
  journal={arXiv preprint arXiv:2311.04400},
  year={2023}
}

@inproceedings{huang2020arch,
  title={Arch: Animatable reconstruction of clothed humans},
  author={Huang, Zeng and Xu, Yuanlu and Lassner, Christoph and Li, Hao and Tung, Tony},
  booktitle={Proceedings of the IEEE/CVF Conference on Computer Vision and Pattern Recognition},
  pages={3093--3102},
  year={2020}
}

@inproceedings{he2021arch++,
  title={Arch++: Animation-ready clothed human reconstruction revisited},
  author={He, Tong and Xu, Yuanlu and Saito, Shunsuke and Soatto, Stefano and Tung, Tony},
  booktitle={Proceedings of the IEEE/CVF international conference on computer vision},
  pages={11046--11056},
  year={2021}
}

@article{zheng2021pamir,
  title={Pamir: Parametric model-conditioned implicit representation for image-based human reconstruction},
  author={Zheng, Zerong and Yu, Tao and Liu, Yebin and Dai, Qionghai},
  journal={IEEE transactions on pattern analysis and machine intelligence},
  volume={44},
  number={6},
  pages={3170--3184},
  year={2021},
  publisher={IEEE}
}

@inproceedings{hu2024animate,
  title={Animate anyone: Consistent and controllable image-to-video synthesis for character animation},
  author={Hu, Li},
  booktitle={Proceedings of the IEEE/CVF Conference on Computer Vision and Pattern Recognition},
  pages={8153--8163},
  year={2024}
}

@article{zhu2024champ,
  title={Champ: Controllable and consistent human image animation with 3d parametric guidance},
  author={Zhu, Shenhao and Chen, Junming Leo and Dai, Zuozhuo and Xu, Yinghui and Cao, Xun and Yao, Yao and Zhu, Hao and Zhu, Siyu},
  journal={arXiv preprint arXiv:2403.14781},
  year={2024}
}

@inproceedings{xu2024magicanimate,
  title={Magicanimate: Temporally consistent human image animation using diffusion model},
  author={Xu, Zhongcong and Zhang, Jianfeng and Liew, Jun Hao and Yan, Hanshu and Liu, Jia-Wei and Zhang, Chenxu and Feng, Jiashi and Shou, Mike Zheng},
  booktitle={Proceedings of the IEEE/CVF Conference on Computer Vision and Pattern Recognition},
  pages={1481--1490},
  year={2024}
}

@inproceedings{smplx:2019,
  title={Expressive body capture: 3d hands, face, and body from a single image},
  author={Pavlakos, Georgios and Choutas, Vasileios and Ghorbani, Nima and Bolkart, Timo and Osman, Ahmed AA and Tzionas, Dimitrios and Black, Michael J},
  booktitle={CVPR},
  year={2019}
}

@article{loper2015smpl,
  title={SMPL: a skinned multi-person linear model},
  author={Loper, Matthew and Mahmood, Naureen and Romero, Javier and Pons-Moll, Gerard and Black, Michael J},
  journal={TOG},
  volume={34},
  number={6},
  pages={1--16},
  year={2015},
  publisher={ACM New York, NY, USA}
}

@misc{MHR:2025,
      title={MHR: Momentum Human Rig},
      author={Aaron Ferguson and Ahmed A. A. Osman and Berta Bescos and Carsten Stoll and Chris Twigg and Christoph Lassner and David Otte and Eric Vignola and Fabian Prada and Federica Bogo and Igor Santesteban and Javier Romero and Jenna Zarate and Jeongseok Lee and Jinhyung Park and Jinlong Yang and John Doublestein and Kishore Venkateshan and Kris Kitani and Ladislav Kavan and Marco Dal Farra and Matthew Hu and Matthew Cioffi and Michael Fabris and Michael Ranieri and Mohammad Modarres and Petr Kadlecek and Rawal Khirodkar and Rinat Abdrashitov and Romain Prévost and Roman Rajbhandari and Ronald Mallet and Russell Pearsall and Sandy Kao and Sanjeev Kumar and Scott Parrish and Shoou-I Yu and Shunsuke Saito and Takaaki Shiratori and Te-Li Wang and Tony Tung and Yichen Xu and Yuan Dong and Yuhua Chen and Yuanlu Xu and Yuting Ye and Zhongshi Jiang},
      year={2025},
      eprint={2511.15586},
      archivePrefix={arXiv},
      primaryClass={cs.GR},
      url={https://arxiv.org/abs/2511.15586},
}

@Article{kerbl3Dgaussians,
      author       = {Kerbl, Bernhard and Kopanas, Georgios and Leimk{\"u}hler, Thomas and Drettakis, George},
      title        = {3D Gaussian Splatting for Real-Time Radiance Field Rendering},
      journal      = {TOG},
      year         = {2023},
      url          = {https://repo-sam.inria.fr/fungraph/3d-gaussian-splatting/}
}

@inproceedings{zheng2024gpsgaussian,
  title={GPS-Gaussian: Generalizable Pixel-wise 3D Gaussian Splatting for Real-time Human Novel View Synthesis},
  author={Zheng, Shunyuan and Zhou, Boyao and Shao, Ruizhi and Liu, Boning and Zhang, Shengping and Nie, Liqiang and Liu, Yebin},
  booktitle={Proceedings of the IEEE/CVF Conference on Computer Vision and Pattern Recognition (CVPR)},
  year={2024}
}

@inproceedings{khirodkar2024sapiens,
  title={Sapiens: Foundation for human vision models},
  author={Khirodkar, Rawal and Bagautdinov, Timur and Martinez, Julieta and Zhaoen, Su and James, Austin and Selednik, Peter and Anderson, Stuart and Saito, Shunsuke},
  booktitle={European Conference on Computer Vision},
  pages={206--228},
  year={2024},
  organization={Springer}
}

@inproceedings{mildenhall2020nerf,
  title={{NeRF}: Representing scenes as neural radiance fields for view synthesis},
  author={Mildenhall, Ben and Srinivasan, Pratul P and Tancik, Matthew and Barron, Jonathan T and Ramamoorthi, Ravi and Ng, Ren},
  booktitle=ECCV,
  pages={405--421},
  year={2020},
}

@misc{chen2024meshavatar,
    title={MeshAvatar: Learning High-quality Triangular Human Avatars from Multi-view Videos}, 
    author={Yushuo Chen and Zerong Zheng and Zhe Li and Chao Xu and Yebin Liu},
    year={2024},
    eprint={2407.08414},
    archivePrefix={arXiv},
    primaryClass={cs.CV},
    url={https://arxiv.org/abs/2407.08414}, 
}

@inproceedings{li2024animatable,
  title={Animatable gaussians: Learning pose-dependent gaussian maps for high-fidelity human avatar modeling},
  author={Li, Zhe and Zheng, Zerong and Wang, Lizhen and Liu, Yebin},
  booktitle={Proceedings of the IEEE/CVF Conference on Computer Vision and Pattern Recognition},
  pages={19711--19722},
  year={2024}
}

@inproceedings{peng2021animatable,
  title={Animatable neural radiance fields for modeling dynamic human bodies},
  author={Peng, Sida and Dong, Junting and Wang, Qianqian and Zhang, Shangzhan and Shuai, Qing and Zhou, Xiaowei and Bao, Hujun},
  booktitle={Proceedings of the IEEE/CVF International Conference on Computer Vision},
  pages={14314--14323},
  year={2021}
}

@inproceedings{weng2022humannerf,
  title={Humannerf: Free-viewpoint rendering of moving people from monocular video},
  author={Weng, Chung-Yi and Curless, Brian and Srinivasan, Pratul P and Barron, Jonathan T and Kemelmacher-Shlizerman, Ira},
  booktitle={CVPR},
  year={2022}
}

@inproceedings{tao2021function4d,
title={Function4D: Real-time Human Volumetric Capture from Very Sparse Consumer RGBD Sensors},
author={Yu, Tao and Zheng, Zerong and Guo, Kaiwen and Liu, Pengpeng and Dai, Qionghai and Liu, Yebin},
booktitle={IEEE/CVF Conference on Computer Vision and Pattern Recognition},
month={June},
year={2021}
}

@article{shao2024human4dit,
title={Human4DiT: 360-degree Human Video Generation with 4D Diffusion Transformer},
author={Shao, Ruizhi and Pang, Youxin and Zheng, Zerong and Sun, Jingxiang and Liu, Yebin},
journal={TOG},
volume={43},
number={6},
articleno={},
year={2024}, publisher={ACM New York, NY, USA}
}

@inproceedings{esser2024scaling,
  title={Scaling rectified flow transformers for high-resolution image synthesis},
  author={Esser, Patrick and Kulal, Sumith and Blattmann, Andreas and Entezari, Rahim and M{\"u}ller, Jonas and Saini, Harry and Levi, Yam and Lorenz, Dominik and Sauer, Axel and Boesel, Frederic and others},
  booktitle={ICML},
  year={2024}
}

@article{zhuang2024idol,
  title={IDOL: Instant Photorealistic 3D Human Creation from a Single Image},
  author={Zhuang, Yiyu and Lv, Jiaxi and Wen, Hao and Shuai, Qing and Zeng, Ailing and Zhu, Hao and Chen, Shifeng and Yang, Yujiu and Cao, Xun and Liu, Wei},
  journal={arXiv preprint arXiv:2412.14963},
  year={2024}
}

@article{li2024pshuman,
  title={PSHuman: Photorealistic Single-view Human Reconstruction using Cross-Scale Diffusion},
  author={Li, Peng and Zheng, Wangguandong and Liu, Yuan and Yu, Tao and Li, Yangguang and Qi, Xingqun and Li, Mengfei and Chi, Xiaowei and Xia, Siyu and Xue, Wei and others},
  journal={arXiv preprint arXiv:2409.10141},
  year={2024}
}

@inproceedings{moon2024exavatar,
  title={Expressive Whole-Body 3D Gaussian Avatar},
  author = {Moon, Gyeongsik and Shiratori, Takaaki and Saito, Shunsuke},  
  booktitle={ECCV},
  year={2024}
}

@inproceedings{li2024animatablegaussians,
  title={Animatable Gaussians: Learning Pose-dependent Gaussian Maps for High-fidelity Human Avatar Modeling},
  author={Li, Zhe and Zheng, Zerong and Wang, Lizhen and Liu, Yebin},
  booktitle={CVPR},
  year={2024}
}

@inproceedings{qiu2024AniGS,
  title={AniGS: Animatable Gaussian Avatar from a Single Image with Inconsistent Gaussian Reconstruction},
  author={Qiu, Lingteng and Zhu, Shenhao and Zuo, Qi and Gu, Xiaodong and Dong, Yuan and Zhang, Junfei and Xu, Chao and Li, Zhe and Yuan, Weihao and Bo, Liefeng and others},
  booktitle={CVPR},
  year={2025}
}

@inproceedings{qiu2025lhm,
  title={LHM: Large Animatable Human Reconstruction Model for Single Image to 3D in Seconds},
  author={Qiu, Lingteng and Gu, Xiaodong and Li, Peihao and Zuo, Qi and Shen, Weichao and Zhang, Junfei and Qiu, Kejie and Yuan, Weihao and Chen, Guanying and Dong, Zilong and others},
  booktitle={Proceedings of the IEEE/CVF International Conference on Computer Vision},
  pages={14184--14194},
  year={2025}
}

@article{qiu2025pflhm,
  title={PF-LHM: 3D Animatable Avatar Reconstruction from Pose-free Articulated Human Images},
  author={Qiu, Lingteng and Li, Peihao and Zuo, Qi and Gu, Xiaodong and Dong, Yuan and Yuan, Weihao and Zhu, Siyu and Han, Xiaoguang and Chen, Guanying and Dong, Zilong},
  journal={arXiv preprint arXiv:2506.13766},
  year={2025}
}

@article{cheng2025wananimate,
  title={Wan-Animate: Unified Character Animation and Replacement with Holistic Replication},
  author={Cheng, Gang and Gao, Xin and Hu, Li and Hu, Siqi and Huang, Mingyang and Ji, Chaonan and Li, Ju and Meng, Dechao and Qi, Jinwei and Qiao, Penchong and others},
  journal={arXiv preprint arXiv:2509.14055},
  year={2025}
}

@inproceedings{guo2023vid2avatar,
  title={Vid2avatar: 3d avatar reconstruction from videos in the wild via self-supervised scene decomposition},
  author={Guo, Chen and Jiang, Tianjian and Chen, Xu and Song, Jie and Hilliges, Otmar},
  booktitle={Proceedings of the IEEE/CVF Conference on Computer Vision and Pattern Recognition},
  pages={12858--12868},
  year={2023}
}

@inproceedings{guo2025vid2avatar,
  title={Vid2avatar-pro: Authentic avatar from videos in the wild via universal prior},
  author={Guo, Chen and Li, Junxuan and Kant, Yash and Sheikh, Yaser and Saito, Shunsuke and Cao, Chen},
  booktitle={Proceedings of the Computer Vision and Pattern Recognition Conference},
  pages={5559--5570},
  year={2025}
}

@article{cai2025up2you,
  title={UP2You: Fast Reconstruction of Yourself from Unconstrained Photo Collections},
  author={Cai, Zeyu and Li, Ziyang and Li, Xiaoben and Li, Boqian and Wang, Zeyu and Zhang, Zhenyu and Xiu, Yuliang},
  journal={arXiv preprint arXiv:2509.24817},
  year={2025}
}

@article{chen2025synchuman,
  title={SyncHuman: Synchronizing 2D and 3D Generative Models for Single-view Human Reconstruction},
  author={Chen, Wenyue and Li, Peng and Zheng, Wangguandong and Zhao, Chengfeng and Li, Mengfei and Zhu, Yaolong and Dou, Zhiyang and Wang, Ronggang and Liu, Yuan},
  journal={arXiv preprint arXiv:2510.07723},
  year={2025}
}

@inproceedings{zheng2023learning,
  title={Learning visibility field for detailed 3D human reconstruction and relighting},
  author={Zheng, Ruichen and Li, Peng and Wang, Haoqian and Yu, Tao},
  booktitle={Proceedings of the IEEE/CVF Conference on Computer Vision and Pattern Recognition},
  pages={216--226},
  year={2023}
}

@article{yang2026sam3dbody,
  title={SAM 3D Body: Robust Full-Body Human Mesh Recovery},
  author={Yang, Xitong and Kukreja, Devansh and Pinkus, Don and Sagar, Anushka and Fan, Taosha and Park, Jinhyung and Shin, Soyong and Cao, Jinkun and Liu, Jiawei and Ugrinovic, Nicolas and Feiszli, Matt and Malik, Jitendra and Dollar, Piotr and Kitani, Kris},
  journal={arXiv preprint arXiv:2602.15989},
  year={2026}
}

@inproceedings{jiang2022neuman,
  title={NeuMan: Neural Human Radiance Field from a Single Video},
  author={Jiang, Wei and Yi, Kwang Moo and Samei, Golnoosh and Tuzel, Oncel and Ranjan, Anurag},
  booktitle={Proceedings of the European conference on computer vision (ECCV)},
  year={2022}
}

@InProceedings{Kanazawa_2019_CVPR,
author = {Kanazawa, Angjoo and Zhang, Jason Y. and Felsen, Panna and Malik, Jitendra},
title = {Learning 3D Human Dynamics From Video},
booktitle = {Proceedings of the IEEE/CVF Conference on Computer Vision and Pattern Recognition (CVPR)},
month = {June},
year = {2019}
}

@inproceedings{du2023agrol,
  author    = {Du, Yuming and Kips, Robin and Pumarola, Albert and Starke, Sebastian and Thabet, Ali and Sanakoyeu, Artsiom},
  title     = {Avatars Grow Legs: Generating Smooth Human Motion from Sparse Tracking Inputs with Diffusion Model},
  booktitle   = {CVPR},
  year      = {2023},
}

@misc{lei2023gart,
      title={GART: Gaussian Articulated Template Models}, 
      author={Jiahui Lei and Yufu Wang and Georgios Pavlakos and Lingjie Liu and Kostas Daniilidis},
      year={2023},
      eprint={2311.16099},
      archivePrefix={arXiv},
      primaryClass={cs.CV}
}

@inproceedings{peng2021neural,
  title={Neural body: Implicit neural representations with structured latent codes for novel view synthesis of dynamic humans},
  author={Peng, Sida and Zhang, Yuanqing and Xu, Yinghao and Wang, Qianqian and Shuai, Qing and Bao, Hujun and Zhou, Xiaowei},
  booktitle={Proceedings of the IEEE/CVF Conference on Computer Vision and Pattern Recognition},
  pages={9054--9063},
  year={2021}
}

@inproceedings{alldieck2022phorhum,
  title	  = {Photorealistic Monocular 3D Reconstruction of Humans Wearing Clothing},
  author  = {Thiemo Alldieck and Mihai Zanfir and Cristian Sminchisescu},
  year	  = {2022},
  booktitle = {Proceedings of the IEEE/CVF Conference on Computer Vision and Pattern Recognition (CVPR)}
}

@inproceedings{SMPL-X:2019,
  title = {Expressive Body Capture: {3D} Hands, Face, and Body from a Single Image},
  author = {Pavlakos, Georgios and Choutas, Vasileios and Ghorbani, Nima and Bolkart, Timo and Osman, Ahmed A. A. and Tzionas, Dimitrios and Black, Michael J.},
  booktitle = {Proceedings IEEE Conf. on Computer Vision and Pattern Recognition (CVPR)},
  pages     = {10975--10985},
  year = {2019}
}

@inproceedings{yariv2021volume,
	  title={Volume rendering of neural implicit surfaces},
	  author={Yariv, Lior and Gu, Jiatao and Kasten, Yoni and Lipman, Yaron},
	  booktitle={Advances in Neural Information Processing Systems},
	  year={2021}
	}

@inproceedings{ARAH:ECCV:2022,
  title = {ARAH: Animatable Volume Rendering of Articulated Human SDFs},
  author = {Shaofei Wang and Katja Schwarz and Andreas Geiger and Siyu Tang},
  booktitle = {European Conference on Computer Vision (ECCV)},
  year = {2022}
}

@article{
	habermann2021,
	author = {Marc Habermann and Lingjie Liu and Weipeng Xu and Michael Zollhoefer and Gerard Pons-Moll and Christian Theobalt},
	title = {Real-time Deep Dynamic Characters},
	journal = {ACM Transactions on Graphics}, 
	month = {aug},
	volume = {40},
	number = {4}, 
	articleno = {94},
	year = {2021}, 
	publisher = {ACM}
}

@inproceedings{moreau2024human,
  title={Human gaussian splatting: Real-time rendering of animatable avatars},
  author={Moreau, Arthur and Song, Jifei and Dhamo, Helisa and Shaw, Richard and Zhou, Yiren and P{\'e}rez-Pellitero, Eduardo},
  booktitle={CVPR},
  year={2024}
}

@InProceedings{Pang_2024_CVPR,
   author    = {Pang, Haokai and Zhu, Heming and Kortylewski, Adam and Theobalt, Christian and Habermann, Marc},
   title     = {ASH: Animatable Gaussian Splats for Efficient and Photoreal Human Rendering},
   booktitle = {Proceedings of the IEEE/CVF Conference on Computer Vision and Pattern Recognition (CVPR)},
   month     = {June},
   year      = {2024},
   pages     = {1165-1175}
}

@misc{jung2023deformable3dgaussiansplatting,
      title={Deformable 3D Gaussian Splatting for Animatable Human Avatars}, 
      author={HyunJun Jung and Nikolas Brasch and Jifei Song and Eduardo Perez-Pellitero and Yiren Zhou and Zhihao Li and Nassir Navab and Benjamin Busam},
      year={2023},
      eprint={2312.15059},
      archivePrefix={arXiv},
      primaryClass={cs.CV},
      url={https://arxiv.org/abs/2312.15059}, 
}

@misc{li2024gaussianbodyclothedhumanreconstruction,
      title={GaussianBody: Clothed Human Reconstruction via 3d Gaussian Splatting}, 
      author={Mengtian Li and Shengxiang Yao and Zhifeng Xie and Keyu Chen},
      year={2024},
      eprint={2401.09720},
      archivePrefix={arXiv},
      primaryClass={cs.CV},
      url={https://arxiv.org/abs/2401.09720}, 
}

@inproceedings{xu2022sanerf,
    author    = {Xu, Tianhan and Fujita, Yasuhiro and Matsumoto, Eiichi},
    title     = {Surface-Aligned Neural Radiance Fields for Controllable 3D Human Synthesis},
    booktitle = {CVPR},
    year      = {2022},
}

@inproceedings{HVTR:3DV2022,
      title={HVTR: Hybrid Volumetric-Textural Rendering for Human Avatars},
      author={Hu, Tao and Yu, Tao and Zheng, Zerong and Zhang, He and Liu, Yebin and Zwicker, Matthias},
      booktitle = {2022 International Conference on 3D Vision (3DV)},
      year = {2022}
}

@inproceedings{2021narf,
  author    = {Noguchi, Atsuhiro and Sun, Xiao and Lin, Stephen and Harada, Tatsuya},
  title     = {Neural Articulated Radiance Field},
  booktitle = {International Conference on Computer Vision},
  year      = {2021},
}

@inproceedings{li2022tava,
  title={TAVA: Template-free animatable volumetric actors},
  author={Li, Ruilong and Tanke, Julian and Vo, Minh and Zollhofer, Michael and Gall, Jurgen and Kanazawa, Angjoo and Lassner, Christoph},
  booktitle={European Conference on Computer Vision (ECCV)},
  year={2022}
}

@inproceedings{zhang2021stnerf,
            title={Editable Free-Viewpoint Video using a Layered Neural Representation},
            author={Jiakai, Zhang
                    and Xinhang, Liu
                    and Xinyi, Ye
                    and Fuqiang, Zhao
                    and Yanshun, Zhang
                    and Minye, Wu
                    and Yingliang, Zhang
                    and Lan, Xu
                    and Jingyi, Yu
                    },
            year={2021},
            booktitle={ACM SIGGRAPH},
           }

@article{zheng2023avatarrex,
author={Zheng, Zerong and Zhao, Xiaochen and Zhang, Hongwen and Liu, Boning and Liu, Yebin},
title={AvatarRex: Real-time Expressive Full-body Avatars},
journal={ACM Transactions on Graphics (TOG)},
volume={42},
number={4},
articleno={},
year={2023}
}

@article{li2023posevocab,
author={Li, Zhe and Zheng, Zerong and Liu, Yuxiao and Zhou, Boyao and Liu, Yebin},
title={PoseVocab: Learning Joint-structured Pose Embeddings for Human Avatar Modeling},
journal={ACM SIGGRAPH Conference Proceedings},
year={2023}
}

@InProceedings{kwon2024ghg,
  title={Generalizable Human Gaussians for Sparse View Synthesis},
  author={Youngjoong Kwon and Baole Fang and Yixing Lu and Haoye Dong and Cheng Zhang and Francisco Vicente Carrasco and Albert Mosella-Montoro and Jianjin Xu and Shingo Takagi and Daeil Kim and Aayush Prakash and Fernando De la Torre},
  booktitle={European Conference on Computer Vision},
  year={2024}
}

@InProceedings{Zhao_2022_CVPR,
    author    = {Zhao, Fuqiang and Yang, Wei and Zhang, Jiakai and Lin, Pei and Zhang, Yingliang and Yu, Jingyi and Xu, Lan},
    title     = {HumanNeRF: Efficiently Generated Human Radiance Field From Sparse Inputs},
    booktitle = {Proceedings of the IEEE/CVF Conference on Computer Vision and Pattern Recognition (CVPR)},
    month     = {June},
    year      = {2022},
    pages     = {7743-7753}
}

@inproceedings{yu2025humanram,
  title={Humanram: Feed-forward human reconstruction and animation model using transformers},
  author={Yu, Zhiyuan and Li, Zhe and Bao, Hujun and Yang, Can and Zhou, Xiaowei},
  booktitle={Proceedings of the Special Interest Group on Computer Graphics and Interactive Techniques Conference Conference Papers},
  pages={1--13},
  year={2025}
}

@InProceedings{Chen_2023_CVPR,
    author    = {Chen, Jianchuan and Yi, Wentao and Ma, Liqian and Jia, Xu and Lu, Huchuan},
    title     = {GM-NeRF: Learning Generalizable Model-Based Neural Radiance Fields From Multi-View Images},
    booktitle = {Proceedings of the IEEE/CVF Conference on Computer Vision and Pattern Recognition (CVPR)},
    month     = {June},
    year      = {2023},
    pages     = {20648-20658}
}

@inproceedings{chen2022gpnerf,
	title={Geometry-guided progressive NeRF for generalizable and efficient neural human rendering},
	author={Chen, Mingfei and Zhang, Jianfeng and Xu, Xiangyu and Liu, Lijuan and Cai, Yujun and Feng, Jiashi and Yan, Shuicheng},
	booktitle={ECCV},
	year={2022}
}

@article{liu2021neural,
      title={Neural Actor: Neural Free-view Synthesis of Human Actors with Pose Control}, 
      author={Lingjie Liu and Marc Habermann and Viktor Rudnev and Kripasindhu Sarkar and Jiatao Gu and Christian Theobalt},
      year={2021},
      journal = {ACM Trans. Graph.(ACM SIGGRAPH Asia)}
}

@misc{zielonka2023drivable3dgaussianavatars,
      title={Drivable 3D Gaussian Avatars}, 
      author={Wojciech Zielonka and Timur Bagautdinov and Shunsuke Saito and Michael Zollhöfer and Justus Thies and Javier Romero},
      year={2023},
      eprint={2311.08581},
      archivePrefix={arXiv},
      primaryClass={cs.CV},
      url={https://arxiv.org/abs/2311.08581}, 
}

@inproceedings{remelli2022drivable,
  title={Drivable volumetric avatars using texel-aligned features},
  author={Remelli, Edoardo and Bagautdinov, Timur and Saito, Shunsuke and Wu, Chenglei and Simon, Tomas and Wei, Shih-En and Guo, Kaiwen and Cao, Zhe and Prada, Fabian and Saragih, Jason and others},
  booktitle={ACM SIGGRAPH 2022 Conference Proceedings},
  pages={1--9},
  year={2022}
}

@article{10.1145/3697140,
	author = {Zhu, Heming and Zhan, Fangneng and Theobalt, Christian and Habermann, Marc},
	title = {TriHuman: A Real-time and Controllable Tri-plane Representation for Detailed Human Geometry and Appearance Synthesis},
	year = {2024},
	publisher = {Association for Computing Machinery},
	address = {New York, NY, USA},
	issn = {0730-0301},
	url = {https://doi.org/10.1145/3697140},
	doi = {10.1145/3697140},
	journal = {ACM Trans. Graph.},
	month = sep
}

@inproceedings{saito2024rgca,
  author = {Shunsuke Saito and Gabriel Schwartz and Tomas Simon and Junxuan Li and Giljoo Nam},
  title = {Relightable Gaussian Codec Avatars}, 
  booktitle = {CVPR},
  year = {2024},
}

@inproceedings{li2024uravatar,
  author = {Junxuan Li and Chen Cao and Gabriel Schwartz and Rawal Khirodkar and Christian Richardt and Tomas Simon and Yaser Sheikh and Shunsuke Saito},
  title = {URAvatar: Universal Relightable Gaussian Codec Avatars}, 
  booktitle = {ACM SIGGRAPH 2024 Conference Papers},
  year = {2024},
}

@article{bagautdinov2021driving,
  title={Driving-signal aware full-body avatars},
  author={Bagautdinov, Timur and Wu, Chenglei and Simon, Tomas and Prada, Fabi{\'a}n and Shiratori, Takaaki and Wei, Shih-En and Xu, Weipeng and Sheikh, Yaser and Saragih, Jason},
  journal={ACM Transactions on Graphics (TOG)},
  volume={40},
  number={4},
  pages={1--17},
  year={2021},
  publisher={ACM New York, NY, USA}
}

@inproceedings{chen2025taoavatar,
  title={Taoavatar: Real-time lifelike full-body talking avatars for augmented reality via 3d gaussian splatting},
  author={Chen, Jianchuan and Hu, Jingchuan and Wang, Gaige and Jiang, Zhonghua and Zhou, Tiansong and Chen, Zhiwen and Lv, Chengfei},
  booktitle={Proceedings of the Computer Vision and Pattern Recognition Conference},
  pages={10723--10734},
  year={2025}
}

@inproceedings{shen2023xavatar,
      title={X-Avatar: Expressive Human Avatars},
      author={Shen, Kaiyue and Guo, Chen and Kaufmann, Manuel and Zarate, Juan and Valentin, Julien and Song, Jie and Hilliges, Otmar},    
      booktitle   = {Computer Vision and Pattern Recognition (CVPR)},
      year      = {2023}
    }

@misc{simeoni2025dinov3,
  title={{DINOv3}},
  author={Sim{\'e}oni, Oriane and Vo, Huy V. and Seitzer, Maximilian and Baldassarre, Federico and Oquab, Maxime and Jose, Cijo and Khalidov, Vasil and Szafraniec, Marc and Yi, Seungeun and Ramamonjisoa, Micha{\"e}l and Massa, Francisco and Haziza, Daniel and Wehrstedt, Luca and Wang, Jianyuan and Darcet, Timoth{\'e}e and Moutakanni, Th{\'e}o and Sentana, Leonel and Roberts, Claire and Vedaldi, Andrea and Tolan, Jamie and Brandt, John and Couprie, Camille and Mairal, Julien and J{\'e}gou, Herv{\'e} and Labatut, Patrick and Bojanowski, Piotr},
  year={2025},
  eprint={2508.10104},
  archivePrefix={arXiv},
  primaryClass={cs.CV},
  url={https://arxiv.org/abs/2508.10104},
}

@inproceedings{chan2019everybody,
  title={Everybody dance now},
  author={Chan, Caroline and Ginosar, Shiry and Zhou, Tinghui and Efros, Alexei A},
  booktitle={Proceedings of the IEEE/CVF international conference on computer vision},
  pages={5933--5942},
  year={2019}
}

@inproceedings{liu2019liquid,
  title={Liquid warping gan: A unified framework for human motion imitation, appearance transfer and novel view synthesis},
  author={Liu, Wen and Piao, Zhixin and Min, Jie and Luo, Wenhan and Ma, Lin and Gao, Shenghua},
  booktitle={Proceedings of the IEEE/CVF international conference on computer vision},
  pages={5904--5913},
  year={2019}
}

@inproceedings{zhao2022thin,
  title={Thin-plate spline motion model for image animation},
  author={Zhao, Jian and Zhang, Hui},
  booktitle={Proceedings of the IEEE/CVF conference on computer vision and pattern recognition},
  pages={3657--3666},
  year={2022}
}

@inproceedings{wang2024disco,
  title={Disco: Disentangled control for realistic human dance generation},
  author={Wang, Tan and Li, Linjie and Lin, Kevin and Zhai, Yuanhao and Lin, Chung-Ching and Yang, Zhengyuan and Zhang, Hanwang and Liu, Zicheng and Wang, Lijuan},
  booktitle={Proceedings of the IEEE/CVF Conference on Computer Vision and Pattern Recognition},
  pages={9326--9336},
  year={2024}
}

@article{zhang2024mimicmotion,
  title={Mimicmotion: High-quality human motion video generation with confidence-aware pose guidance},
  author={Zhang, Yuang and Gu, Jiaxi and Wang, Li-Wen and Wang, Han and Cheng, Junqi and Zhu, Yuefeng and Zou, Fangyuan},
  journal={arXiv preprint arXiv:2406.19680},
  year={2024}
}

@article{martinez2024codec,
  author = {Julieta Martinez and Emily Kim and Javier Romero and Timur Bagautdinov and Shunsuke Saito and Shoou-I Yu and Stuart Anderson and Michael Zollh{\"o}fer and Te-Li Wang and Shaojie Bai and Chenghui Li and Shih-En Wei and Rohan Joshi and Wyatt Borsos and Tomas Simon and Jason Saragih and Paul Theodosis and Alexander Greene and Anjani Josyula and Silvio Mano Maeta and Andrew I. Jewett and Simon Venshtain and Christopher Heilman and Yueh-Tung Chen and Sidi Fu and Mohamed Ezzeldin A. Elshaer and Tingfang Du and Longhua Wu and Shen-Chi Chen and Kai Kang and Michael Wu and Youssef Emad and Steven Longay and Ashley Brewer and Hitesh Shah and James Booth and Taylor Koska and Kayla Haidle and Matt Andromalos and Joanna Hsu and Thomas Dauer and Peter Selednik and Tim Godisart and Scott Ardisson and Matthew Cipperly and Ben Humberston and Lon Farr and Bob Hansen and Peihong Guo and Dave Braun and Steven Krenn and He Wen and Lucas Evans and Natalia Fadeeva and Matthew Stewart and Gabriel Schwartz and Divam Gupta and Gyeongsik Moon and Kaiwen Guo and Yuan Dong and Yichen Xu and Takaaki Shiratori and Fabian Prada and Bernardo R. Pires and Bo Peng and Julia Buffalini and Autumn Trimble and Kevyn McPhail and Melissa Schoeller and Yaser Sheikh},
  title = {{Codec Avatar Studio: Paired Human Captures for Complete, Driveable, and Generalizable Avatars}},
  year = {2024},
  journal = {NeurIPS Track on Datasets and Benchmarks},
}

@misc{li2026largescalecodecavatarsunreasonable,
      title={Large-scale Codec Avatars: The Unreasonable Effectiveness of Large-scale Avatar Pretraining},
      author={Junxuan Li and Rawal Khirodkar and Chengan He and Zhongshi Jiang and Giljoo Nam and Lingchen Yang and Jihyun Lee and Egor Zakharov and Zhaoen Su and Rinat Abdrashitov and Yuan Dong and Julieta Martinez and Kai Li and Qingyang Tan and Takaaki Shiratori and Matthew Hu and Peihong Guo and Xuhua Huang and Ariyan Zarei and Marco Pesavento and Yichen Xu and He Wen and Teng Deng and Wyatt Borsos and Anjali Thakrar and Jean-Charles Bazin and Carsten Stoll and Ginés Hidalgo and James Booth and Lucy Wang and Xiaowen Ma and Yu Rong and Sairanjith Thalanki and Chen Cao and Christian Häne and Abhishek Kar and Sofien Bouaziz and Jason Saragih and Yaser Sheikh and Shunsuke Saito},
      year={2026},
      eprint={2604.02320},
      archivePrefix={arXiv},
      primaryClass={cs.CV},
      url={https://arxiv.org/abs/2604.02320},
}


\end{document}